\documentclass[journal]{IEEEtran}
\usepackage{amsmath,amssymb,amsfonts}
\usepackage{bm}
\usepackage{bbm}
\usepackage{algorithmic}
\usepackage{algorithm}
\usepackage{array}
\usepackage[font=small]{caption}
\usepackage{textcomp}
\usepackage{stfloats}
\usepackage{multicol}
\usepackage{multirow}
\usepackage{url}
\usepackage{lipsum}
\usepackage{verbatim}
\usepackage{graphicx}
\usepackage{cite}
\usepackage{xcolor}
\usepackage{booktabs}
\usepackage{placeins}
\usepackage{soul}
\setstcolor{red}

\renewcommand\a{\bm{a}}%
\newcommand\q{\bm{q}}

\newcommand\x{\bm{x}}
\newcommand\y{\bm{y}}
\newcommand\z{\bm{z}}

\newcommand\alphab{\boldsymbol{\alpha}}

\newcommand\lambdab{\boldsymbol{\lambda}}
\newcommand\mub{\boldsymbol{\mu}}

\newcommand\A{\bm{A}}

\newcommand\Y{\bm{Y}}
\newcommand\Z{\bm{Z}}

\newcommand\At{\bm{\mathcal{A}}}

\newcommand\Xt{\bm{\mathcal{X}}}

\newcommand\THETA{\bm{\Theta}}

\newcommand\RR{\mathbb{R}}

\definecolor{JungleGreen}{rgb}{0,0.7,0.4}
\definecolor{deepPurple}{rgb}{0.5,0.1,0.5}
\definecolor{bittersweet}{rgb}{1.0, 0.44, 0.37}

\newcommand{\revOne}[1]{\textcolor{black}{#1}}
\newcommand{\revTwo}[1]{\textcolor{black}{#1}}
\newcommand{\revThree}[1]{\textcolor{black}{#1}}
\newcommand{\revFour}[1]{\textcolor{black}{#1}}
\newcommand{\revFive}[1]{\textcolor{black}{#1}}
\newcommand{\revOneB}[1]{\textcolor{black}{#1}}

\newcommand\argmax[1]{\underset{#1}{\arg\,\mathrm{max}}\,}

\newcommand{\rev}[1]{{\color[rgb]{0,0,0} #1}}

\begin{document}

\title{\rev{Joint Bayesian Parameter and Model Order Estimation for Low-Rank Probability Mass Tensors}}

\author{Joseph K. Chege,~\IEEEmembership{Graduate Student Member, IEEE}, Arie Yeredor,~\IEEEmembership{Fellow, IEEE}, and Martin Haardt,~\IEEEmembership{Fellow, IEEE}
\thanks{Joseph K. Chege and Martin Haardt are with the Communications Research Laboratory, Ilmenau University of Technology, Ilmenau, Germany (e-mail: joseph.chege@tu-ilmenau.de, martin.haardt@tu-ilmenau.de).}
\thanks{Arie Yeredor is with the School of Electrical and Computer Engineering, Tel Aviv University, Tel Aviv, Israel (e-mail: arie@eng.tau.ac.il).}
\thanks{The authors gratefully acknowledge the support of the German Research Foundation (DFG) under the PROMETHEUS project (reference no. HA 2239/16-1, project no. 462458843).}}



\maketitle

\begin{abstract}
Obtaining a reliable estimate of the joint probability mass function (PMF) of a set of random variables from observed data is a significant objective in statistical signal processing and machine learning. 
Modelling the joint PMF as a tensor that admits a low-rank canonical polyadic decomposition (CPD) has enabled the development of efficient PMF estimation algorithms.
However, these algorithms require the rank (model order) of the tensor to be specified beforehand.
In real-world applications, the true rank is unknown. 
Therefore, an appropriate rank is usually selected from a candidate set either by observing validation errors or by computing various likelihood-based information criteria, a procedure \rev{that could be costly in terms of computational time or hardware resources, or \revFour{could} result in mismatched models which affect \revFour{the} model accuracy}.
This paper presents a novel Bayesian framework for estimating the \rev{low-rank components of a} joint PMF \rev{tensor} and \rev{simultaneously} inferring its rank from \revFour{the} observed data.
We specify a Bayesian PMF estimation model and employ appropriate prior distributions for the model parameters, allowing \rev{the rank to be inferred without cross-validation}.
We then derive a deterministic solution based on variational inference (VI) to approximate the posterior distributions of various model parameters. 
%
%
Numerical experiments involving both synthetic data and real \rev{classification and item} recommendation data illustrate the advantages of our \revTwo{VI-based} method in terms of estimation accuracy, automatic rank detection, and computational efficiency.
\end{abstract}


\section{Introduction} \label{intro}
In statistical signal processing and machine learning, one of the most essential, yet challenging, problems is obtaining a reliable estimate of the joint distribution of a set of random variables from observed data.
In the case of multivariate discrete random variables, the joint distribution is represented by a joint probability mass function (PMF) which describes the probability of each of the possible realizations of the random variables.
Knowledge of the joint PMF enables many statistical learning tasks to be solved in a\revTwo{n explainable} and optimal manner.
For instance, two common tasks which are ubiquitous in statistical learning are predicting missing data (e.g., item ratings) given a subset of observations\rev{,}~and predicting class \rev{or} cluster labels given the associated features\rev{\cite{vapnik2013nature}}. 
Applications of these learning tasks arise in recommender systems (e.g., for movies, music, etc.) and data classification \rev{or} clustering, respectively\rev{\cite{melville2011recommender, bishop2006}}.
Using the joint PMF, the posterior distribution of the missing data given the observed data can be computed, from which the conditional expectation of missing data can be found. 
Similarly, the posterior distribution of the class labels given the features can be computed from the joint PMF. The predicted class label is then the maximizer of the posterior distribution.
It is known that these estimates are optimal in that they minimize the mean squared error and the probability of misclassification, respectively (see, e.g., \cite{bishop2006}).

A classical nonparametric technique for estimating a joint PMF from observed data is histogram estimation, i.e., counting the \revFive{relative} number of instances each possible realization of the random variables is observed. 
However, this approach faces the curse of dimensionality: the number of observations required for a reliable estimate grows exponentially with the number of variables. 
This drawback limits the applicability of histogram estimation for many practical cases of interest.
Therefore, recent work on PMF estimation has focused on developing alternative approaches which mitigate the curse of dimensionality. 
These approaches exploit the fact that a joint PMF of a set of $N$ random variables can \revFive{often} be represented as an $N$-dimensional tensor which admits a low-rank canonical polyadic decomposition (CPD) (see, e.g., \cite{kolda_tensor_2009}, \cite{sidiropoulos_tensor_2017}). 
Importantly, the work in \cite{ishteva_tensors_2015} showed the connection between various latent variable models and tensor decompositions. In particular, the author drew a parallel between the na\"{i}ve Bayes model and the CPD.
This connection was proved more rigorously in \cite{kargas_tensors_2018}, where it was shown that any joint PMF can be represented by a na\"{i}ve Bayes model with one latent variable taking a finite number of states. 
Furthermore, the number of latent states is the rank of the joint PMF tensor, i.e., the minimum number of rank-one components in the CPD.
\rev{In addition}, \rev{t}he authors \rev{proved that the joint PMF of all the variables can be recovered from lower-order (e.g., third-order) marginal PMF tensors.} 
Further works have proposed various PMF estimation approaches, ranging from coupled tensor factorization based on lower-order marginals \cite{amiridi_statistical_2019, yeredor_estimation_2019, ibrahim_recovering_2021, flores_coupled_2022} to maximum-likelihood based approaches in which no lower-order marginals need to be computed \cite{yeredor_maximum_2019, chege_efficient_2022}.

In all the aforementioned approaches, the rank (or the model order) of the joint PMF tensor needs to be somehow specified before estimation.
However, while there exist some results regarding bounds on the tensor rank (see, e.g., \cite{kolda_tensor_2009}, \cite{sidiropoulos_tensor_2017}), in real-world scenarios, the exact tensor rank is unknown. 
In practice, the most common approach is to treat the rank as a hyperparameter to be tuned via cross-validation (CV) (this is done, e.g., in \cite{kargas_tensors_2018}). 
Alternatively, there are \rev{classical} model order selection criteria based on likelihood functions.
These include the Akaike information criterion (AIC) \cite{akaike1974new} and the Bayesian information criterion (BIC) \cite{schwarz1978estimating}. 
In recent years, the decomposed \revTwo{normalized} maximum likelihood criterion (DNML) \cite{yamanishi_decomposed_2019}, which is based on the minimum description length principle, has also been proposed for model order selection in hierarchical latent variable models.
While these techniques are well-established and have theoretical guarantees, they require that the rank be selected from a prespecified candidate set.
%
\rev{If the training cost of the estimation algorithm is high (e.g., for datasets with many samples or random variables), such a} model order selection procedure is costly in terms of computation time \rev{or hardware resources due to the requirement to carry out multiple training runs}. 
One way to \rev{reduce} the computation cost could be to decrease the size of the candidate set, e.g., by \rev{limiting} the range of \revFive{tested orders} or by \revFive{testing orders on a sparse grid}.
However, there is always the possibility of model order misspecification in case the true model order is not included in the candidate set.  
%

%
%
%
%

\rev{The objective of this paper is to develop an efficient method to estimate the parameters of a low-rank joint PMF tensor as well as its model order, or rank. Our contributions are summarized as follows:}
\begin{itemize}
    \item \rev{We propose a novel and efficient Bayesian framework to estimate both the low-rank joint PMF tensor of a set of discrete random variables and its rank in a unified manner. Our method enables the rank of the PMF tensor to be estimated as part of the Bayesian inference procedure without resorting to cross-validation or other model order selection techniques as in existing PMF tensor estimation approaches (Section~\ref{sec:pmf}).}

    \item \rev{The connection between the na\"{i}ve Bayes model and the CPD of a joint PMF tensor is exploited to specify a Bayesian model, tailored to low-rank PMF estimation, that ensures \revFour{that} the estimated parameters satisfy the probability simplex (nonnegativity and sum-to-one) constraints. A sparsity-promoting Dirichlet prior is imposed on the loading vector (the weights of the latent components) to enable identification and pruning of irrelevant components without heuristic thresholding or manual model selection (Subsection~\ref{subsec:model_spec}).}

    \item \rev{We develop a PMF estimation algorithm based on variational inference (VI) to approximate posterior distributions over the CPD model parameters. The VI approach provides closed-form updates for all posterior distributions, enabling efficient implementation and deterministic convergence without sampling. Moreover, we derive a pruning threshold for irrelevant components that depends on the prior hyperparameter and the dataset size, avoiding \revFive{\it{ad hoc}} threshold specification (Subsections~\ref{subsec:VB}, \ref{subsec:vb_pmf}).}

    \item \rev{We validate the accuracy and rank estimation performance of our proposed method through carefully designed synthetic data simulations, which include varying dataset sizes, PMF tensor ranks, outage probabilities (fraction of missing data), and robustness to the choice of \revFour{the} hyperparameter for the loading vector prior (Section~\ref{sec:synthetic}). Moreover, we present results using real datasets to demonstrate applicability to practical classification and item recommendation tasks (Section~\ref{sec:real}).}
\end{itemize}


\revTwo{An initial version of the VI-based} approach was presented at CAMSAP 2023 \cite{chege_bayesian_2023}. 
\revOne{Here, we provide more details on the Bayesian model specification and the derivation of important expressions, an in-depth discussion on the rank inference properties of the algorithm, as well as an appendix elaborating how the optimization objective function is derived.} \rev{We also present \revFive{additional} synthetic data simulations under a variety of scenarios, as well as a set of real-data recommendation and classification experiments to further validate our approach.}
\subsection{\rev{Related Work}}
\rev{This paper is primarily concerned with developing a method for joint parameter and model order estimation for PMF tensors. As such, our work is closely related to existing CPD-based PMF tensor estimation approaches~\cite{kargas_tensors_2018, yeredor_estimation_2019, amiridi_statistical_2019, ibrahim_recovering_2021, flores_coupled_2022, yeredor_maximum_2019, chege_efficient_2022}. Unlike these works, our method directly estimates the model order (rank) of the PMF tensor from the data, avoiding manual rank selection. Moreover, we take a Bayesian approach tailored to low-rank PMF tensor estimation, and no computation of lower-order marginals is required as in \cite{kargas_tensors_2018, yeredor_estimation_2019, amiridi_statistical_2019, ibrahim_recovering_2021, flores_coupled_2022}.
\revOneB{In recent years, tensor networks have been used to model joint probability distributions for supervised and unsupervised learning \cite{glasser2019expressive, glasser2020probabilistic, novikov2021tensor, liu2023tensor}. In particular, \cite{glasser2019expressive} discusses several tensor-network (TN) representations of joint PMF tensors and compares their expressive capacities (i.e., the set of functions that such networks can efficiently represent). The authors also propose the locally purified state (LPS) representation, which performs better than the other TN variants considered, although the model order (the purification rank) is selected manually.}  

Nonparametric Bayesian approaches have been used for \revFour{the} probabilistic modeling of categorical random variables \cite{dunson_nonparametric_2009, bhattacharya2012simplex, zhou_bayesian_2015, yang2016bayesian}. These works use infinite mixtures based on stick-breaking Dirichlet process priors and apply Markov chain Monte Carlo (MCMC) sampling methods for posterior inference.
%
%
%
In contrast, \revFour{we use a finite mixture model}, based on simple Dirichlet priors, whose fixed latent structure is a nonnegative CPD which admits a na\"{i}ve Bayes model interpretation, allowing \revFour{a} universal representation of any joint PMF as well as uniqueness under mild conditions \cite{kargas_tensors_2018}. Although the models in \cite{dunson_nonparametric_2009} and \cite{zhou_bayesian_2015} can be viewed as a CPD, the connection to the na\"{i}ve Bayes model \revFive{becomes loose} due to the use of infinite mixtures. 
On the other hand, the models in \cite{bhattacharya2012simplex} and \cite{yang2016bayesian} admit a higher-order singular value decomposition (HOSVD), which is not unique \cite{kolda_tensor_2009}. Moreover, \revFour{we adopt \revOneB{VI}} which yields closed-form \revOneB{variational} updates \revOneB{and avoids explicit posterior}
sampling\revOneB{. While MCMC techniques provide asymptotically exact posterior inference through sampling from the target distribution, they are often computationally expensive in practice}~\cite{blei_variational_2017}, \revOneB{and require careful assessment to determine convergence~\cite{gelman_bayesian_2013}}. \revOneB{Our variational Bayesian framework thus provides a computationally efficient alternative to MCMC inference, at the cost of replacing asymptotically exact sampling-based inference with an approximate variational solution.} 

The CPD representation of a joint PMF tensor\revFive{,} along with the na\"{i}ve Bayes interpretation\revFive{,} results in a latent variable model that, at first glance, \revFive{resembles} \revFour{the mixture-of-unigrams} model used in document analysis \cite{nigam2000text, blei2003latent}. \revFour{The mixture-of-unigrams model describes the distribution of words in a single document by selecting one topic (the latent class) and sampling all word tokens independently from that topic, making it a latent class model for repeated observations of a single variable (the document). This differs fundamentally from our model, which directly parametrizes and estimates the full joint PMF of \textit{multiple} discrete variables.} Moreover, our approach also estimates the model order along with the model parameters, whereas the aforementioned works focus on estimating the model parameters.

The automatic rank estimation property of our method relies on a sparsity-promoting Dirichlet prior distribution. Thus, our work can be considered part of the general framework of sparse Bayesian learning, in which Bayesian shrinkage priors play an important role (see, e.g., \cite{cheng_rethinking_2022, vandecappelle_inexact_2021} for an overview). Such shrinkage priors have been used for rank estimation in Bayesian tensor decomposition models \cite{zhao_bayesian_2015, cheng_probabilistic_2017, cheng_learning_2020, cheng_towards_2022}, where the goal is to decompose multidimensional data to discover underlying structures and patterns. Dirichlet process priors have been used in nonparametric Bayesian tensor decomposition models, e.g., \cite{porteous2008multi, zhe2015scalable}. In contrast, we use finite Dirichlet priors and our goal is to model and estimate the joint PMF tensor of a set of discrete random variables given a (two-dimensional) dataset of observations. Furthermore, our choice of the Dirichlet prior is informed by the fact that the commonly used Bayesian shrinkage priors (such as the Gaussian scale mixture family \cite{cheng_rethinking_2022}) are not suitable for PMF tensor estimation since they do not naturally fulfill the probability simplex constraints.}   

\subsection{Notation}
This paper uses bold lowercase letters (e.g., $\a$) to denote vectors, bold uppercase letters (e.g., $\A$) to represent matrices, and bold calligraphic letters (e.g., $\At$) to represent tensors.
Uppercase letters (e.g., $A$) denote scalar random variables while lowercase letters (e.g. $a$) denote realizations of the corresponding random variables. 
The outer product is represented by $\circ$ while $^\mathsf{T}$ denotes the transpose operator. \rev{The squared Frobenius norm is denoted by the operator $\|\cdot\|_\mathsf{F}^2$.}
\section{Problem Statement} \label{sec:prel}
Consider a discrete random vector $\x = [X_1, \dotsc, X_N]^\mathsf{T} \in \RR ^ N$ where each random variable $X_n$ can take discrete values in $\{1, \dotsc, I_n\}$, $n = 1, \dotsc, N$. 
The joint PMF $p(\x)$ can be described by an $N$-dimensional tensor $\Xt \in \RR ^ {I_1 \times \dots \times I_N}$ whereby each element of the tensor is the joint probability of a particular realization of the random variables, i.e., $\mathsf{Pr}(X_1 = i_1, \dotsc, X_N = i_N) = \Xt(i_1, \dotsc, i_N)$.
In general, $\Xt$ can be expressed as a sum of rank-one \revFive{nonnegative} terms via the \revFive{(nonnegative)} CPD
\begin{equation} \label{eq:cpd}
\Xt = \sum_{r=1}^R \lambda_r \A_1(:, r) \circ \A_2(:, r) \dotsc \circ \A_N(:, r),
\end{equation}
where $\A_n \in \RR ^ {I_n \times R}$ denotes a factor matrix \revFive{with nonnegative elements,} and $\A_n(:,r)$ denotes the $r$-th column of $\A_n$. 
Without loss of generality, the columns of the factor matrices can be restricted to have unit norm such that $\|\A_n(:,r)\|_p = 1$ for $p \ge 1$, $\forall ~ n,r$. Then, $\lambdab = [\lambda_1,\dotsc, \lambda_r]^\mathsf{T}$ is the loading vector\revFive{, consisting of positive elements,} which `absorbs' the norms of the columns. The minimum number $R$ of rank-1 terms required for the decomposition to hold is referred to as the rank\footnote{\revFive{or the positive rank\revFour{\cite{lim2009nonnegative}}}} of the tensor. In terms of the CPD, a particular element of the tensor is given by
\begin{equation} \label{eq:tensor_element}
    \Xt(i_1,\dotsc,i_N) = \sum_{r=1}^R \lambda_r \prod_{n=1}^N \A_n(i_n,r).
\end{equation}

It has been shown in \cite{ishteva_tensors_2015} and \cite{kargas_tensors_2018} that a \revFive{(nonnegative)} rank-$R$ CPD can be interpreted as a particular kind of latent variable model known as the na\"{i}ve Bayes model. This model assumes that the random variables $\{X_n\}_{n=1}^N$ are conditionally independent given a hidden (latent) variable $H$ which can take a finite number of $R$ states. Under this interpretation, each element $\lambda_r$ of the loading vector is the probability $\mathsf{Pr}(H=r)$ of the hidden variable, while each factor matrix column $\A_n(:,r)$ \revFour{represents} the conditional PMF $p(X_n \, | \, H = r)$. Thus, \eqref{eq:cpd} is subject to a set of probability simplex constraints, i.e., $\lambdab > \bm{0}, \A_n \ge \bm{0}$ (nonnegativity) and $\bm{1} ^ \mathsf{T} \lambdab = 1$, $\bm{1} ^ \mathsf{T} \A_n = \bm{1} ^ \mathsf{T}$ (sum-to-one).

We assume that we observe a discrete random vector $\y = [Y_1,\dotsc,Y_N]^\mathsf{T}$ according to the model (see, e.g., \cite{yeredor_maximum_2019})
\begin{equation} \label{eq:model}
Y_n
=
\begin{cases}
X_n & \text{w.p. } 1-p \\
0 & \text{w.p. } p
\end{cases}
,\quad n=1,\dotsc,N,
\end{equation}
(independently for each $n$ and independently of the values in $\x$), where the outage probability $p$ denotes the probability that $X_n$ is unobserved in $\y$. We have at hand a dataset $\Y = [\y_1, \dotsc, \y_T]$ containing $T$ i.i.d. realizations of $\y$, where $\y_t = [y_{1,t}, \dotsc, y_{N,t}] ^ \mathsf{T}$. The goal is to estimate the \rev{CPD components $\{\{\A_n\}_{n=1}^N, \lambdab\}$ of the unknown} joint PMF tensor $\Xt$ from the observations $\Y$.

\section{Variational Bayesian PMF Estimation} \label{sec:pmf}
\revFive{In previous work on this topic, e}stimates of the CPD components $\lambdab$ and $\{\A_n\}_{n=1}^N$ (and, therefore, of the joint PMF tensor $\Xt$) \revFive{have been} obtained from $\Y$ by first obtaining empirical estimates of \rev{lower}-order marginal PMF tensors from $\Y$, followed by \revFour{a} coupled tensor factorization (e.g., \cite{kargas_tensors_2018}, \cite{yeredor_estimation_2019}).
Alternatively, a maximum likelihood (ML) estimate of the PMF tensor \revFive{was} obtained by fitting the observed data to the CPD model (e.g., \cite{yeredor_maximum_2019}, \cite{chege_efficient_2022}). While these approaches work quite well, they assume that the CPD rank $R$ is known, a scenario which rarely arises in practice. We therefore seek to formulate the problem within the Bayesian paradigm and design an algorithm to estimate the CPD components \revFive{while automatically detecting} the rank of $\Xt$.

\subsection{Model Specification} \label{subsec:model_spec}
The Bayesian model is specified by first identifying the model parameters tha\revFive{t} one wishes to estimate, finding an expression for the likelihood of the observations given the model parameters, and then assigning prior distributions to the model parameters. 

For the PMF estimation problem, we wish to estimate the CPD components, i.e., the loading vector $\lambdab$ and the columns of the factor matrices $\{\A_n\}_{n=1}^N$. For convenience of notation \revThree{and without loss of generality}, let us assume that $p = 0$, i.e., there are no missing observations in the data. Following the observation model \eqref{eq:model}, the joint likelihood of the observations given the model parameters, up to a constant, is (cf. \eqref{eq:tensor_element})
\begin{equation} \label{eq:likelihood}
   p \big(\Y \, \big| \, \lambdab, \{\A_n\}_{n=1}^N \big) = \prod_{t=1}^{T} \revFive{\Bigg(} \sum_{r=1}^R \lambda_r \prod_{n=1}^N \A_n(y_{n,t}, r)\revFive{\Bigg)}.
\end{equation}
The na\"{i}ve Bayes model \revFive{structure} allows us to introduce a latent variable $\z \in \RR ^ R$ which contains the hidden \revFive{mechanism} that governs the observation $\y$. Let $\z = [z_1, \dotsc, z_R] ^ \mathsf{T}$ where a particular element $z_r$ is equal to 1 and all other elements are equal to 0. Therefore, the elements of $\z$ satisfy $z_r \in \{0,1\}$ and $\sum_{r=1}^R z_r = 1$. Each observation $\y$ is associated with a latent variable $\z$ which identifies the latent component from which $\y$ (and, therefore, $\x$) was drawn. In other words, if $\y$ is drawn from component $s \in \{1,\dotsc, R\}$ then $z_r = 1$ if \revFive{$r = s$} and $z_r = 0$ if \revFive{$r \neq s$}. Defining $\Z = [\z_1,\dotsc,\z_T]$ where $\z_t = [z_{1,t},\dotsc,z_{r,t}]^\mathsf{T}$, we can express the conditional distribution of $\Y$ given $\Z$ and $\{\A_n\}_{n=1}^N$ (up to a constant) as
\begin{equation} \label{eq:cond_Y}
    p \big(\Y \, \big| \, \Z, \{\A_n\}_{n=1}^N \big)  = \prod_{t=1}^T \prod_{r=1}^R \prod_{n=1}^N \Big(\A_n (y_{n,t}, r)\Big) ^ {z_{r,t}}.
\end{equation}

From the definition of the latent variable $\z$, it can be seen that $\mathsf{Pr}(z_r = 1) = \lambda_r$. Due to fact that $\z$ is a one-hot-encoded vector (only one element is equal to 1), we can equivalently express this probability as $p(\z) = \prod_{r=1}^R \lambda_r ^ {z_r}$. Thus, for $T$ observations, the conditional distribution of $\Z$ given $\lambdab$ is\footnote{Note that by multiplying \eqref{eq:cond_Y} and \eqref{eq:p_z_lambda} and marginalizing over $\Z$, we can recover \eqref{eq:likelihood}, i.e., $\sum_{\z_t} \prod_{t=1}^T  p (\y_t \, | \, \z_t, \{\A_n\}_{n=1}^N ) p(\z_t \, | \, \lambdab)=p (\Y \,  | \, \lambdab, \{\A_n\}_{n=1}^N )$, as required.}
\begin{equation} \label{eq:p_z_lambda}
    p(\Z \, | \ \lambdab) = \prod_{t=1}^T \prod_{r=1}^R \lambda_r ^ {z_{r,t}}.
\end{equation}

Next, we specify prior distributions for the component parameters $\lambdab$ and $\{\A_n\}_{n=1}^N$, which are subject to the probability simplex constraints (see Section \ref{sec:prel}). An appropriate distribution for variables defined on a probability simplex is the Dirichlet \revThree{distribution}, which is a multivariate distribution over a set of random variables that are subject to nonnegativity and sum-to-one constraints. In particular, given $K$ random variables $\mub = [\mu_1, \dotsc, \mu_K]^\mathsf{T}$\rev{,} the Dirichlet distribution is given by (e.g., \cite{bishop2006})
\begin{equation} \label{eq:dirichlet}
    \mathrm{Dir}(\mub \, | \, \alphab) = C(\alphab) \prod_{k=1}^K \mu_k ^ {\alpha_k - 1}, ~~\revFive{\mub \in \Delta^{K-1}} ,
\end{equation}
where \revFive{$\Delta^{K-1} \triangleq \big\{\mub \in \RR^K: \mu_k \ge 0, \sum_{k=1}^K \mu_k = 1 \big\}$ is the $(K-1)$-dimensional probability simplex. Furthermore,} $C(\alphab)$ is the normalization constant, defined in terms of the standard gamma function\footnote{\revThree{For any real number $x$, the standard gamma function is defined as $\Gamma(x) = \int_0^\infty t^{x-1} \mathrm{e}^{-t} \mathrm{d}t$.}} $\Gamma(\cdot)$ as
\begin{equation}\label{eq:dir_norm}
    C(\alphab) = \frac{\Gamma \big(\sum_{k=1}^K \alpha_k\big)}{\prod_{k=1}^K \Gamma(\alpha_k)},
\end{equation}
and $\alphab = [\alpha_1, \dotsc, \alpha_K]^\mathsf{T}$ are the Dirichlet parameters\rev{, subject to the constraint $\alpha_k > 0, \forall k$}. These parameters govern how evenly or sparsely distributed the resulting distributions are.  In particular, $\alpha \rightarrow 0$ favors distributions with nearly all mass concentrated on one of their components (i.e., sparse), $\alpha \rightarrow \infty$ favors near-uniform distributions, while for $\alpha=1$, all distributions are equally likely \revFour{(e.g., \cite{bishop2006})}. This property is crucial in designing an algorithm which can automatically detect the model order of the PMF tensor (see Section \ref{subsec:vb_pmf} for more details). 

Some standard properties of the Dirichlet distribution will be useful in the sequel. The \revOne{expected value of a Dirichlet-distributed random variable $\mu_k$} is its normalized \revOne{Dirichlet} parameter, i.e.,
\begin{equation} \label{eq:exp_dir}
    \mathbb{E}[\mu_k] = \frac{\alpha_k}{\sum_{k=1}^K \alpha_k},
\end{equation}
while the expectation of its logarithm is defined in terms of the digamma function $\psi (\cdot)$, which is the first derivative of the logarithm of the standard gamma function, i.e.,
\begin{equation} \label{eq:exp_log_dir}
    \mathbb{E}[\log \mu_k] = \psi(\alpha_k) - \psi \Big(\sum_{k=1}^K \alpha_k\Big).
\end{equation}
where $\psi(x) \triangleq \frac{\mathrm{d}}{\mathrm{d}x} \log \Gamma(x)$.

Having reviewed the Dirichlet distribution, we now apply it to the model parameters. Let $p(\lambdab)$ and $p(\a_{n,r})$ be the prior distributions for the loading vector $\lambdab$ and the $r$-th column $\A_n(:,r)$ of the $n$-th factor matrix $\A_n$, respectively. Further, define $\a_{n,r} \triangleq \A_n(:,r) =  [a_{n,r,1}, \dotsc, a_{n,r,I_n}]^\mathsf{T} \in \RR^{I_n}$. Then,
\begin{align} 
    p(\lambdab) &= \mathrm{Dir}(\lambdab \,|\,\alphab_{\lambda}) = C(\alphab_{\lambda}) \prod_{r = 1} ^ R\lambda_r ^ {\alpha_{\lambda,r}-1} \label{eq:p_lambda}, \\
   p(\a_{n,r})  &= \mathrm{Dir}(\a_{n,r} \, | \, \alphab_{n,r}) = C(\alphab_{n,r}) \prod_{\revFive{i} = 1} ^ {I_n} a_{n,r,\revFive{i}} ^ {\alpha_{n,r,\revFive{i}}-1} \label{eq:p_a},
\end{align}
where $\alphab_{\lambda} = [\alpha_{\lambda,1},\dotsc,\alpha_{\lambda,R}]^\mathsf{T}$ and $\alphab_{n,r} = [\alpha_{n,r,1},\dotsc,\alpha_{n,r,I_n}]^\mathsf{T}$ are the respective parameters for the prior distributions. In the absence of any prior information favoring one element over another, we choose symmetric Dirichlet distributions as priors, i.e., $\alpha_{\lambda,r}=\alpha_{\lambda}\revThree{,} \,\, \forall r$\revThree{,} and  $\alpha_{n,r,\revFive{i}}=\alpha_{n,r}\revThree{,} \,\, \forall \revFive{i}$.

Let $\bm{\Theta} = \{ \Z, \lambdab, \A_1, \dotsc, \A_N\}$ be the collection of all unknown quantities, i.e., latent variables and parameters. The joint distribution of the observations and the unknown quantities is given by
\begin{figure}[t] 
    \centering
    \includegraphics[width=0.5\linewidth]{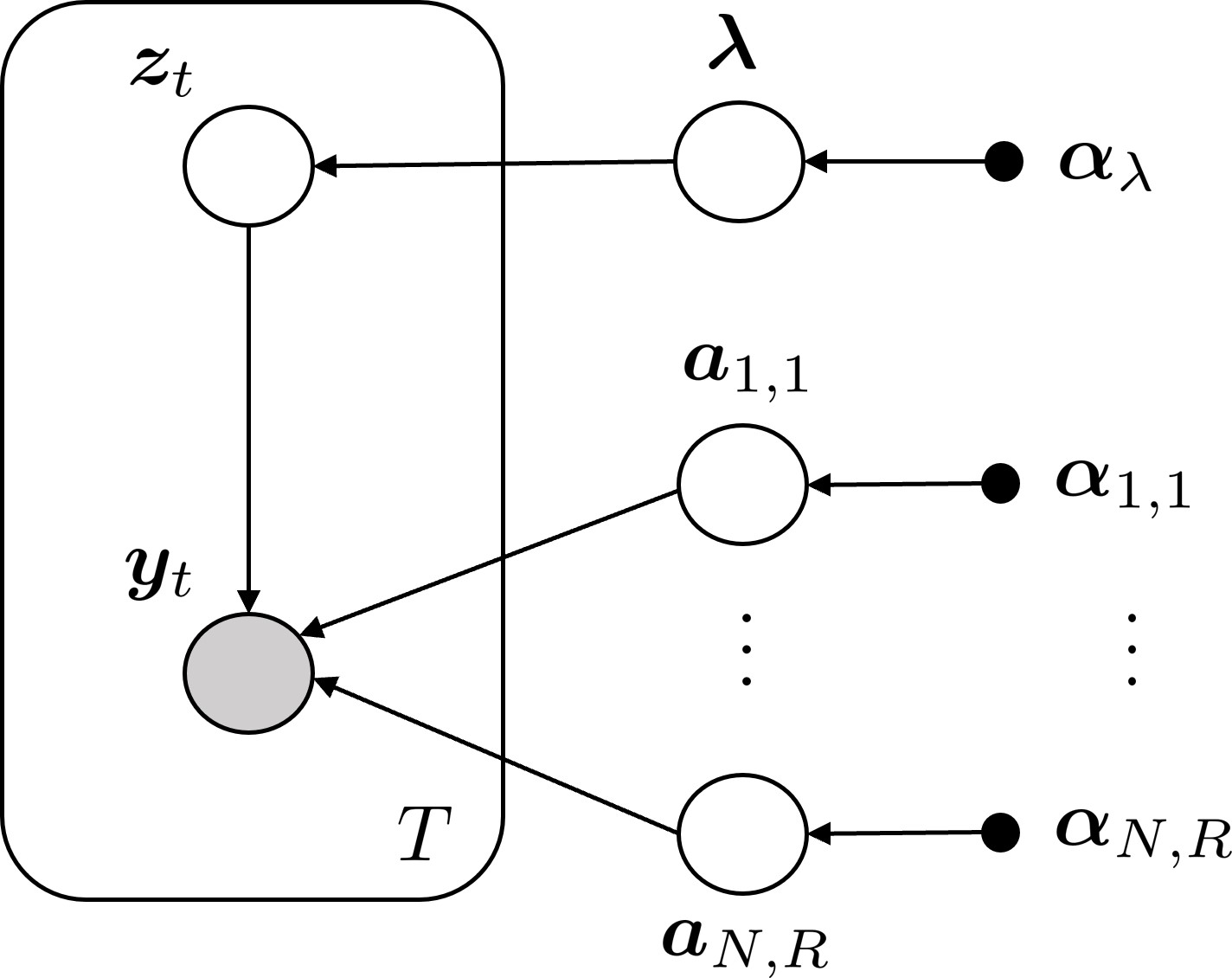}
    \caption{Probabilistic graphical model representation of the joint distribution $p(\Y,\THETA)$. \revFour{The local and global random variables are represented by the larger circles, whereas the hyperparameters are depicted by the smaller solid dots. The shaded circle denotes that $y_t$ is an observed realization. The plate represents $T$ instances, of which only \{$y_t$, $z_t$\} are shown explicitly.}}
    \label{fig:graph_model}
\end{figure}
\begin{equation} \label{eq:joint_distrib}
    \begin{aligned}[b]
        p(\Y,\THETA) &= p \big(\Y \, \big| \, \Z, \{\A_n\}_{n=1}^N \big) \cdot p(\Z \, | \, \lambdab)\\
        &\qquad \times p(\lambdab) \cdot \prod_{n=1}^N \prod_{r=1}^R p(\a_{n,r}).
    \end{aligned}
\end{equation}
The probabilistic graphical model representation of $\eqref{eq:joint_distrib}$ is shown in Fig.~\ref{fig:graph_model}. The latent variables $\{\z_t\}_{t=1}^T$ are said to be \textit{local} parameters since $\z_t$ only governs the data in the $t$-th context. On the other hand, $\lambdab$ and $\a_{n,r}, \forall n,r$\revThree{,} are \revFour{considered} \textit{global} parameters \revFour{because their total number is fixed and does not scale with the dataset}\rev{\cite{blei_variational_2017}}. 
%
\revFour{From \eqref{eq:joint_distrib},} the logarithm of the joint distribution is given by (cf. \eqref{eq:cond_Y}, \eqref{eq:p_z_lambda}, \eqref{eq:p_lambda}, and  \eqref{eq:p_a})
\begin{equation} \label{eq:log_joint}
    \begin{aligned} [b]   
        \log p(\Y, \bm{\Theta}) & = \sum_{t=1}^T  \sum_{r=1}^R z_{r,t} \Big(\log \lambda_r + \sum_{n=1}^N  \log a_{n,r,y_{n,t}} \Big)   \\
        & \hspace{-2em} + (\alpha_{\lambda} - 1) \sum_{r=1}^R\log \lambda_r \\
        & \hspace{-2em} + \sum_{n=1}^N \sum_{r=1}^R \sum_{\revFive{i}=1}^{I_n}(\alpha_{n,r} - 1) \log a_{n,r,\revFive{i}} + \mathrm{const},
    \end{aligned}
\end{equation}
where `const' refers to terms which do not depend on $\Y$ or $\THETA$, \revThree{and $a_{n,r,y_{n,t}} \triangleq \A_n(y_{n,t}, r)$.}

Our goal is to infer the posterior distribution $p(\THETA\,|\,\Y)$, i.e., the distribution of the unknown quantities given the observations. The classical approach is through Bayes theorem,
\begin{equation} \label{eq:bayes}
    p(\THETA\,|\,\Y) = \frac{p(\Y,\THETA)}{\int_{\THETA} p(\Y,\THETA) \mathrm{d}\THETA}.
\end{equation}
Exact inference requires integration over all variables in $\THETA$ to calculate the denominator. In the present case, this is a high-dimensional problem that involves integration over the model parameters and summation over all possible (discrete) states of the latent variables, which is computationally infeasible. In the following, we resort to approximate Bayesian inference, which yields closed-form expressions for the posterior distribution.  


\subsection{Posterior Approximation using Variational Inference} \label{subsec:VB}
\begingroup
\renewcommand{\arraystretch}{2}
\begin{table*}[]
\centering
\caption{\rev{Summary of priors, posteriors, update equations, and point estimates for variational Bayesian PMF estimation.}}
\begin{tabular}{|c|c|c|c|c|}
\hline
\textbf{Parameter}        & \textbf{Prior}                                      & \textbf{Posterior}                                             & \textbf{Update Equations}                                                                                                                                                                                                                  & \textbf{Point Estimate}                                                                                    \\ \hline
$\lambdab \in \RR^R$      & $\mathrm{Dir}(\lambdab \, | \, \alphab_{\lambda})$  & $\mathrm{Dir}(\lambdab \, | \, \widetilde{\alphab}_{\lambda})$ & $\displaystyle \widetilde{\alpha}_{\lambda,r} = \alpha_{\lambda} + \sum_{t=1}^T \rho_{r,t}$                                                                                                                                                    & $\displaystyle \widehat{\lambda}_r = \frac{\widetilde{\alpha}_{\lambda,r}}{\sum_{j=1}^R \widetilde{\alpha}_{\lambda,j}}$ \\ \hline
$\a_{n,r} \in \RR^{I_n}$  & $\mathrm{Dir}(\a_{n,r} \, | \, \alphab_{n,r})$      & $\mathrm{Dir}(\a_{n,r} \, | \, \widetilde{\alphab}_{n,r})$     & $ \displaystyle \widetilde{\alpha}_{n,r,\revFive{i}} = \alpha_{n,r,\revFive{i}} + \sum_{t \in \Omega_{n,i_n}} \rho_{r,t}$                                                                                                                                              & $\displaystyle \widehat{a}_{n,r,\revFive{i}} = \frac{\widetilde{\alpha}_{n,r,\revFive{i}}}{\sum_{j=1}^{I_n} \widetilde{\alpha}_{n,r,j}}$ \\ \hline
$\Z \in \RR^{R \times T}$ & $\displaystyle \prod_{t=1}^T \prod_{r=1}^R \lambda_r ^ {z_{r,t}}$ & $ \displaystyle \prod_{t=1}^T \prod_{r=1}^R \rho_{r,t} ^ {z_{r,t}}$         & \begin{tabular}[c]{@{}c@{}}$\gamma_{r, t} = \exp{\left \{\mathbb{E}[\log \lambda_r] + \sum_{n=1}^N \mathbb{E}[\log a_{n,r,y_{n,t}}]  \right \}}$\\ $\displaystyle \rho_{r,t} = \frac{\gamma_{r,t}}{\sum_{j=1}^R \gamma_{j,t}}$\end{tabular} & $\widehat{z}_{r,t} = \mathbb{E}[z_{r,t}] = \rho_{r,t}$                                                     \\ \hline
\end{tabular}
\label{tab:summary}
\end{table*}
\endgroup
In variational inference (VI) \cite{jordan_introduction_1998, blei_variational_2017}, we seek a variational distribution $q(\bm{\Theta})$ that is closest to the true posterior distribution $p(\bm{\Theta} \, | \, \Y)$ such that the Kullback-Leibler divergence (KLD) between them is minimized. The KLD is given by
\begin{equation} \label{eq:KLD}
    \begin{aligned}[b]
        D(q(\bm{\Theta})\,\| \,p(\bm{\Theta} \,|\,\Y)) & = \int_{\THETA} q(\bm{\Theta}) \log \frac{q(\bm{\Theta})}{p(\bm{\Theta}\,|\,\Y)} \mathrm{d}\bm{\Theta} \\
        & \hspace{-5em}= \log p(\Y) - \underbrace{\int_{\THETA} \q(\bm{\Theta}) \log  \frac{p(\Y, \bm{\Theta})}{q(\bm{\Theta})} \mathrm{d}\bm{\Theta}}_{\mathcal{L}(q)}.
    \end{aligned}
\end{equation}
A closer examination of \eqref{eq:KLD} reveals that since the KLD is nonnegative and the marginal likelihood \rev{$p(\Y) = \int_{\THETA}p(\Y,\THETA)\mathrm{d}\THETA$} is independent of $\bm{\Theta}$, then maximizing the term $\mathcal{L}(q)$ is equivalent to minimizing $D(q(\bm{\Theta})\,\| \,p(\bm{\Theta} \,|\,\Y))$. Therefore, $\mathcal{L}(q)$ is a lower bound on $\log p(\Y)$ and is usually referred to as the evidence lower bound (ELBO). 

If all possible variational distributions $q(\THETA)$ are considered, then the KLD vanishes when $q(\THETA) = p(\THETA\,|\,\Y)$. However, since the true posterior distribution is intractable, some restrictions need to be imposed on $q(\THETA)$ to ensure tractability. A common way to restrict the family of variational distributions is to assume that $q(\bm{\Theta})$ can be factorized such that $q(\THETA) = \prod_i q_i(\THETA_i)$, where $\cup_i \THETA_i = \THETA$ and $\cap_i \THETA_i = \varnothing$. \rev{Therefore, each of the unknown quantities $\THETA_i$ is assumed to be governed by a distinct variational distribution $q_i(\THETA_i)$.}  This approach is called mean-field approximation\rev{\cite{jordan_introduction_1998, blei_variational_2017}}.
In our case, \revThree{given that $\bm{\Theta} = \{ \Z, \lambdab, \A_1, \dotsc, \A_N\}$, the mean-field approximation yields}  
\begin{equation} \label{eq:mean_field}
    q(\bm{\Theta}) = \prod_{t=1}^T q_z(\z_t) \cdot q_{\lambda}(\lambdab) \cdot \prod_{n=1}^N \prod_{r=1}^R q_{n,r}(\a_{n,r}).
\end{equation}

\rev{Under the mean-field approximation, the optimized form of the $i$th factor $q_i ^ {\ast}(\THETA_i)$ is \cite{bishop2006}}
\begin{equation} \label{eq:optimal_dist}
    \log q_i ^ {\ast}(\THETA_i) = \mathbb{E}_{q(\bm{\Theta} \backslash \THETA_i)}[\log p(\Y, \THETA)] + \mathrm{const}\rev{,}
\end{equation}
\rev{which is derived by replacing $q(\THETA)$ in the expression for $\mathcal{L}(q)$ in \eqref{eq:KLD} with its factorized form, fixing all but the $i$th factor, and maximizing $\mathcal{L}(q)$ with respect to $q_i(\THETA_i)$.}
Here, $\mathbb{E}_{q(\bm{\Theta} \backslash \THETA_i)}[\cdot]$ denotes the expectation with respect to ${q(\bm{\THETA})}$ over all components except $\THETA_i$, while $\log p(\Y,\THETA)$ is given in \eqref{eq:log_joint}. The constant is found by normalizing the distribution $q_i ^ {\ast}(\THETA_i)$
\begin{equation}\label{eq:norm_q}
    q_i ^ {\ast}(\THETA_i) = \frac{\exp \big(\mathbb{E}_{q(\bm{\Theta} \backslash \THETA_i)}[\log p(\Y, \THETA)]\big)}{\int_{\THETA_i} \exp\big(\mathbb{E}_{q(\bm{\Theta} \backslash \THETA_i)}[\log p(\Y, \THETA)]\big) \mathrm{d}\THETA_i}.
\end{equation}
However, in most cases, the constant will be readily apparent upon inspection of the resulting optimal distributions. 

It can be seen from \eqref{eq:optimal_dist} that computing a particular variational distribution requires knowledge of the other variational distributions. Therefore, the variational distributions are updated iteratively using a coordinate ascent algorithm (see Section \ref{subsec:vb_pmf}). \rev{Since the ELBO is convex with respect to each factor $q_i(\THETA_i)$, coordinate ascent is guaranteed to converge to a local optimum \cite{bishop2006}.} In the following, we present the derivations of each variational distribution in \eqref{eq:mean_field}. \revFive{To keep the notation uncluttered, we will write $\mathbb{E}[\THETA_i]$ in place of $\mathbb{E}_{q_i^\ast(\THETA_i)}[\THETA_i]$ (i.e., the expectation of the $i$th unknown with respect to its variational distribution) where necessary.}

\subsubsection{Posterior distribution $q^{\ast}_{\lambda}(\lambdab)$ of the loading vector $\lambdab$} \label{subsubsec:lambda}

To derive $q^{\ast}_{\lambda}(\lambdab)$, we first select, from \eqref{eq:log_joint}, terms that depend on $\lambdab$ and then apply \eqref{eq:optimal_dist} by evaluating expectations over all other parameters (w.r.t. their respective $q(\cdot)$) except $\lambdab$. We have
\begin{equation} \label{eq:log_q_lambda}
    \log q^{\ast}_{\lambda}(\lambdab) = \sum_{r=1}^R \Big(\alpha_{\lambda}-1 + \sum_{t=1} ^ T \mathbb{E}[z_{r,t}] \Big) \log \lambda_r + \mathrm{const}.
\end{equation}
Taking exponentials on both sides reveals that
\begin{equation}
    q^{\ast}_{\lambda}(\lambdab) \propto \prod_{r=1}^R \lambda_r ^ {(\alpha_{\lambda} + \sum_{t=1} ^ T \mathbb{E}[z_{r,t}]) - 1},
\end{equation}
which we immediately recognize as a Dirichlet distribution with parameters defined by
\begin{equation} \label{eq:alpha_lambda}
    \widetilde{\alpha}_{\lambda, r} \triangleq \alpha_{\lambda} + \sum_{t=1} ^ T \mathbb{E}[z_{r,t}],~ r=1,\dotsc,R.
\end{equation}
Thus, the optimal variational distribution is given by
\begin{equation} \label{eq:opt_q_lambda}
q_{\lambda} ^ {\ast}(\lambdab) = \mathrm{Dir}(\lambdab \, | \, \widetilde{\alphab}_{\lambda}) = C(\widetilde{\alphab}_{\lambda})\prod_{r=1}^R \lambda_r ^ {\widetilde{\alpha}_{\lambda,r} - 1},
\end{equation}
where $\widetilde{\alphab}_{\lambda} = [\widetilde{\alpha}_{\lambda,1}, \dotsc, \widetilde{\alpha}_{\lambda,R}]^\mathsf{T}$ are the global variational parameters (for the global parameter $\lambdab$) and $C(\widetilde{\alphab}_{\lambda})$ is the normalization constant, which can be computed as in \eqref{eq:dir_norm}.
A point estimate $\widehat{\lambdab}$ can be found by computing the posterior expectation over $q^ {\ast}_{\lambda}(\lambdab)$ (cf. \eqref{eq:exp_dir}), i.e.,
\begin{equation} \label{eq:lambda_est}
    \widehat{\lambda}_r = \mathbb{E}_{q_{\lambda} ^ {\ast}(\lambdab)}[\lambda_r] = \frac{\widetilde{\alpha}_{\lambda,r}}{\sum_{j=1}^R\widetilde{\alpha}_{\lambda,j}},~ r=1,\dotsc,R.
\end{equation}
\subsubsection{Posterior distribution $q^{\ast}_{n,r}(\a_{n,r})$ of the factor matrix columns $\a_{n,r}$} \hfill \\

From \eqref{eq:log_joint}, we select \revFour{the} terms with a dependence on $\a_{n,r}$ and then apply \eqref{eq:optimal_dist}. \rev{Isolating the $(n,r)$-th term, we have}
%
\rev{\begin{equation} \label{eq:log_qnr_1}
\begin{aligned} 
    \log q_{n,r}^{\ast}(\a_{n,r}) & = \sum_{t=1}^T \mathbb{E}[z_{r,t}] \log a_{n,r,y_{n,t}} \\ 
    & \hspace{-2em} + \sum_{\revFive{i}=1}^{I_n} (\alpha_{n,r} - 1) \log a_{n,r,\revFive{i}} + \mathrm{const}.
\end{aligned}
\end{equation}}\rev{T}o simplify \eqref{eq:log_qnr_1}, we \revFour{rewrite the first term using a summation over the index $i$}. Since $y_{n,t} \in \{1, \dotsc, I_n\}$, \eqref{eq:log_qnr_1} can be equivalently expressed as
%
\rev{\begin{equation} \label{eq:log_qnr_2}
\begin{aligned}[b]
    \log q_{n,r}^{\ast}(\a_{n,r}) & =  \sum_{\revFive{i} = 1}^{I_n} \Bigg(\Big(\sum_{t \in \Omega_{n,\revFive{i}}}\mathbb{E}[z_{r,t}]\Big) \log a_{n,r,\revFive{i}}\\
    & + (\alpha_{n,r} - 1) \log a_{n,r,\revFive{i}}\Bigg) + \mathrm{const},
\end{aligned}
\end{equation}}where $\Omega_{n,\revFive{i}} \triangleq \{t:y_{n,t}=\revFive{i}\} \subset \{1,\dotsc,T\}$ is the set of indices $t$ in which the observation $y_{n,t}$ equals the discrete value $\revFive{i}$. Restricting the summation range of the index $t$ in this way ensures that \eqref{eq:log_qnr_2} and \eqref{eq:log_qnr_1} are equivalent.
\rev{The optimal variational distribution $q^{\ast}_{n,r}(\a_{n,r})$ is therefore given by}
\begin{equation} \label{eq:qnr}
    q^{\ast}_{n,r}(\a_{n,r}) \propto \prod_{\revFive{i}=1}^{I_n} a_{n,r,\revFive{i}} ^ {(\alpha_{n,r} +\sum_{t \in \Omega_{n,\revFive{i}}}\mathbb{E}[z_{r,t}]) - 1}, ~~ \forall n,r.
\end{equation}
This is also a Dirichlet distribution with parameters given by
\begin{equation} \label{eq:alpha_a}
    \widetilde{\alpha}_{n,r,\revFive{i}} \triangleq \alpha_{n,r} + \sum_{t \in \Omega_{n,\revFive{i}}} \mathbb{E}[z_{r,t}],~\revFive{i} = 1,\dotsc,I_n.
\end{equation}
Thus, the optimal variational distribution is given by
\begin{equation} \label{eq:opt_q_nr}
    q^{\ast}_{n,r}(\a_{n,r}) = \mathrm{Dir}(\a_{n,r}\,|\,\alphab_{n,r}) = C(\widetilde{\alphab}_{n,r})\prod_{\revFive{i}=1}^{I_n}a_{n,r,\revFive{i}}^{\widetilde{\alpha}_{n,r,\revFive{i}}-1},
\end{equation}
where $\widetilde{\alphab}_{n,r} = [\widetilde{\alpha}_{n,r,1},\dotsc,\widetilde{\alpha}_{n,r,I_n}]^\mathsf{T}$\revThree{,} $\forall n,r$\revThree{,} are the global variational parameters (for the global parameters $\a_{n,r}$) and $C(\widetilde{\alphab}_{n,r})$ is the normalization constant, which can be computed as in \eqref{eq:dir_norm}. A point estimate $\widehat{\a}_{n,r}$ is found by computing the posterior expectation over $q^{\ast}_{n,r}(\a_{n,r})$ (cf. \eqref{eq:exp_dir}), i.e.,
\begin{equation} \label{eq:a_est}
    \widehat{a}_{n,r,\revFive{i}} = \mathbb{E}_{q^{\ast}_{n,r}(\a_{n,r})}[a_{n,r,\revFive{i}}] = \frac{\widetilde{\alpha}_{n,r,\revFive{i}}}{\sum_{j=1}^{I_n}\widetilde{\alpha}_{n,r,j}},~\forall \revFive{i}.
\end{equation}

\subsubsection{Posterior distribution $q^{\ast}_z(\z_t)$ of the latent variable $\z_t$}

From \eqref{eq:log_joint}, we select terms that have a dependence on $\z_t$ and then apply \eqref{eq:optimal_dist}. We have
\begin{equation} \label{eq:log_qz}
    \log q^{\ast}_z(\z_t) 
    = \sum_{r=1}^R z_{r,t} \Big(\revFour{\mathbb{E}[\log \lambda_r]} + \sum_{n=1}^N \revFour{\mathbb{E}[\log a_{n,r,y_{n,t}}]} \Big)
\end{equation}
where 
the expectations are taken with respect to the variational distributions $q^\ast_{\lambda}(\lambdab)$ and $q^\ast_{n,r}(\a_{n,r})$, respectively. Therefore, the expectations are given by (cf. \eqref{eq:exp_log_dir})
\begin{equation} \label{eq:expectations}
    \begin{aligned}
        \revFour{\mathbb{E}[\log\lambda_r]}
        & = \psi(\widetilde{\alpha}_{\lambda,r}) - \psi \Big(\sum_{j=1}^R \widetilde{\alpha}_{\lambda,r}\Big), \\
        \revFour{\mathbb{E}[\log a_{n,r,y_{n,t}}]}
        & = \psi(\widetilde{\alpha}_{n,r,y_{n,t}}) - \psi \Big(\sum_{j=1}^{I_n} \widetilde{\alpha}_{n,r,j}\Big). 
    \end{aligned}
\end{equation}
Taking exponentials on both sides of \eqref{eq:log_qz} yields
\begin{equation} \label{eq:q_z_1}
    q^{\ast}_z(\z_t) \propto \prod_{r=1}^R \gamma_{r,t} ^ {z_{r,t}},
\end{equation}
where we have defined $\gamma_{r,t}$ as
\begin{equation} \label{eq:gamma_rt}
    \gamma_{r,t} \triangleq \exp \Big\{\revFour{\mathbb{E}[\log\lambda_r]} + \sum_{n=1}^N  \revFour{\mathbb{E}[\log a_{n,r,y_{n,t}}]} \Big\}.
\end{equation}
Normalizing \eqref{eq:q_z_1} gives us the optimal variational distribution
\begin{equation} \label{eq:q_z_2}
    q^{\ast}_z(\z_t) = \prod_{r=1}^R \rho_{r,t} ^ {z_{r,t}},
\end{equation}
where
\begin{equation} \label{eq:rho_rt}
    \rho_{r,t} \triangleq \frac{\gamma_{r,t}}{\sum_{j=1}^R \gamma_{j,t}}
\end{equation}
is the local variational parameter (for the local parameter $z_{r,t}$).
Recall that $\z_t$ is one-hot-encoded (see Section \ref{subsec:model_spec}). Thus, taking the expectation \revThree{with respect to} \eqref{eq:q_z_2} reveals that
\begin{equation} \label{eq:z_est}
    \widehat{z}_{r,t} = \mathbb{E}_{q^{\ast}_z(\z_t)}[z_{r,t}] = \rho_{r,t},
\end{equation}
and this expectation can be \revFive{substituted} into \eqref{eq:alpha_lambda} and \eqref{eq:alpha_a} to compute the respective Dirichlet parameters.

\rev{Table \ref{tab:summary} summarizes the prior and posterior distributions, update equations, and point estimates for each model parameter.} \revFive{Note that the updates are easily adapted to datasets with missing observations (i.e., $y_{n,t} = 0$; cf. \eqref{eq:model}) by only considering sample indices $t \in \Omega_{n,i}$, i.e., those indices for which $y_{n,t} = i$, where $i=1,\dotsc,I_n$.}

\subsection{Computing the ELBO} \label{subsec:elbo}

As mentioned in Section \ref{subsec:VB}, the goal of variational inference is to find a distribution $q(\THETA)$ that is closest to the true posterior distribution $p(\THETA \,|\,\Y)$ in terms of the KLD. According to \eqref{eq:KLD}, rather than directly optimizing the KLD, we only need to maximize the ELBO $\mathcal{L}(q)$. Since $\mathcal{L}(q)$ is a lower bound, it should increase at each iteration during optimization and can therefore be used to evaluate the correctness of the mathematical expressions and to test for convergence. From \eqref{eq:KLD}, the ELBO can be written as
\begin{equation} \label{eq:elbo_1}
    \mathcal{L}(q) = \mathbb{E}_{q(\THETA)}[\log p(\Y,\THETA)] - \mathbb{E}_{q(\THETA)}[\log q(\THETA)],
\end{equation}
where the first term denotes the posterior expectation of the joint distribution (cf. \eqref{eq:log_joint}), while the second term is the (negative) entropy of the posterior distributions. The expectations are taken with respect to the optimal variational distributions in turn. For the first term in \eqref{eq:elbo_1}, the joint distribution $p(\Y,\THETA)$ can be decomposed into various terms, as shown in \eqref{eq:joint_distrib}. Evaluating the expectations yields the closed-form expression in \eqref{eq:elbo_2} for computing the ELBO. Details on the derivation of \eqref{eq:elbo_2} can be found in the appendix.

\subsection{Algorithm: VB-PMF} \label{subsec:vb_pmf}
%
\begin{figure*}[b]
    \hrule
    \vspace{0.5em}
\begin{equation} \label{eq:elbo_2}
    \begin{aligned}
        \mathcal{L}(q) & =  \sum_{t=1}^T \sum_{r=1}^R \rho_{r,t}  
        \Big(
        \revFour{\mathbb{E}[\log \lambda_r]}
        + \sum_{n=1}^N 
        \revFour{\mathbb{E}[\log a_{n,r,y_{n,t}}]}
        \Big) 
         + \sum_{n=1}^N \sum_{r=1}^R \Big( \log C(\alphab_{n,r}) 
          + (\alpha_{n,r}-1)   \sum_{\revFive{i}=1}^{I_n}
        \revFour{\mathbb{E}[\log a_{n,r,\revFive{i}}]}
          \Big) \\  
       & 
       +  \Big(\log C(\alphab_{\lambda}) + (\alpha_{\lambda} - 1)\sum_{r=1}^R 
       \revFour{\mathbb{E}[\log \lambda_r]}
       \Big) 
       - \sum_{t=1}^T \sum_{r=1}^R \rho_{r,t} \log \rho_{r,t} 
       - \Big(\log C(\widetilde{\alphab}_{\lambda}) + \sum_{r=1}^R (\widetilde{\alpha}_{\lambda,r}-1)
       \revFour{\mathbb{E}[\log \lambda_r]}
       \Big)  \\
       &  - \sum_{n=1}^N \sum_{r=1}^R\Big(\log C(\widetilde{\alphab}_{n,r}) + \sum_{\revFive{i}=1}^{I_n}  (\widetilde{\alpha}_{n,r,\revFive{i}}-1) 
       \revFour{\mathbb{E}[\log a_{n,r,\revFive{i}}]}
       \Big)
    \end{aligned} \tag{39}
\end{equation}
\end{figure*}
%
%
\begin{table}[t]
\normalsize
\begin{tabular}{l}
\hline
\hspace{-0.5em}\textbf{Algorithm: VB-PMF}\\ \hline
\hspace{-0.5em}\textbf{Input:} Dataset $\Y$, initial rank $R$ \\
\hspace{-0.5em}\textbf{Output:} CPD estimates $\widehat{\lambdab}$ and $\{\widehat{\A}_n\}_{n=1}^N$, rank estimate $\widehat{R}$ \\
\hspace{-0.5em}~1:~Set prior hyperparameters $\alpha_{\lambda}$ and $\alpha_{n,r}, \, \forall n,r$. \\
\hspace{-0.5em}~2:~Initialize $\widetilde{\alphab}_{\lambda}$ and $\widetilde{\alphab}_{n,r}, \forall n,r$ randomly.\\
\hspace{-0.5em}~3:~\textbf{repeat} \\
\hspace{-0.5em}~4:~$\quad$ Update $\rho_{r,t}, \, \forall r,t$ using \eqref{eq:expectations}, \eqref{eq:gamma_rt}, and \eqref{eq:rho_rt}.\\
\hspace{-0.5em}~5:~$\quad$ Update $\widetilde{\alpha}_{\lambda,r}, \, \forall r$ using \eqref{eq:alpha_lambda}. \\
\hspace{-0.5em}~6:~$\quad$ Update $\widetilde{\alpha}_{n,r, \revFive{i}}, \, \forall n,r,\revFive{i}$ using \eqref{eq:alpha_a}.  \\
\hspace{-0.5em}~7:~$\quad$ Evaluate the ELBO using \eqref{eq:elbo_2}.  \\
\hspace{-0.5em}~8:~\textbf{until} the ELBO converges  \\
\hspace{-0.5em}~9:~Find $\widehat{\lambdab}$ and $\{\widehat{\A}_n\}_{n=1}^N$ using \eqref{eq:lambda_est} and \eqref{eq:a_est}, respectively.\\
\hspace{-0.6em}10:~Find $\widehat{R}$ by \rev{selecting} components corresponding to \\
$ \quad \widehat{\lambda}_r \rev{> \alpha_{\lambda}/T}$. \\
\hline
\end{tabular}
\end{table}

The variational Bayesian PMF estimation procedure (VB-PMF) is outlined in Algorithm 1. The algorithm estimates the CPD components of the PMF tensor and also simultaneously and automatically infers the tensor rank. The automatic rank detection property arises from exploiting the sparsity-promoting properties of the Dirichlet distribution. Recall that the value of the Dirichlet parameter influences whether the resulting random variables are evenly or sparsely distributed within the probability simplex (see Section \ref{subsec:model_spec}). In particular, if the prior parameter $\alpha_{\lambda}$ is set to a small value close to 0 (e.g., $10^{-6}$), the point estimate $\widehat{\lambdab}$ obtained will be a sparse vector. This can be understood by noticing that \eqref{eq:lambda_est} can be expressed as
\refstepcounter{equation}
\begin{equation} \label{eq:lambda_est_v2}
    \widehat{\lambda}_r = \frac{\alpha_{\lambda} + \sum_{t=1}^T \rho_{r,t}}{R\alpha_{\lambda} + T},
\end{equation}
\begin{figure}[t]
    \hspace{-1em}
    \begin{tabular}{cc}
        \includegraphics[width=0.48\columnwidth]{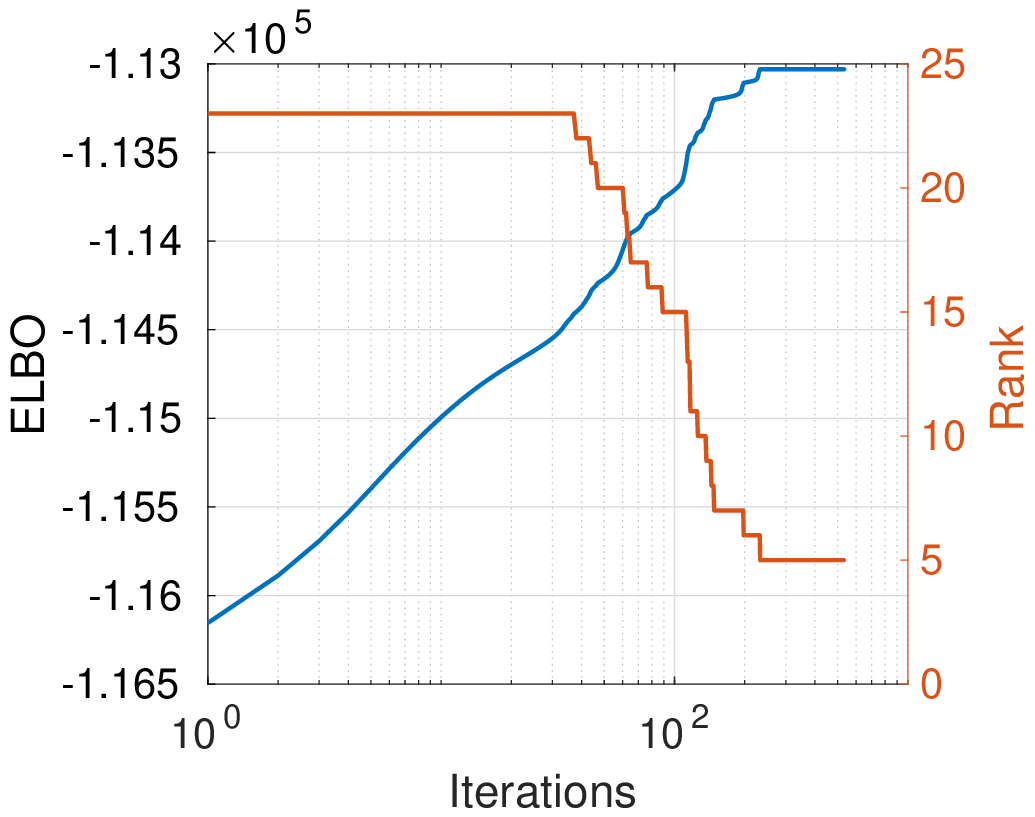} &
        \includegraphics[width=0.48\columnwidth]{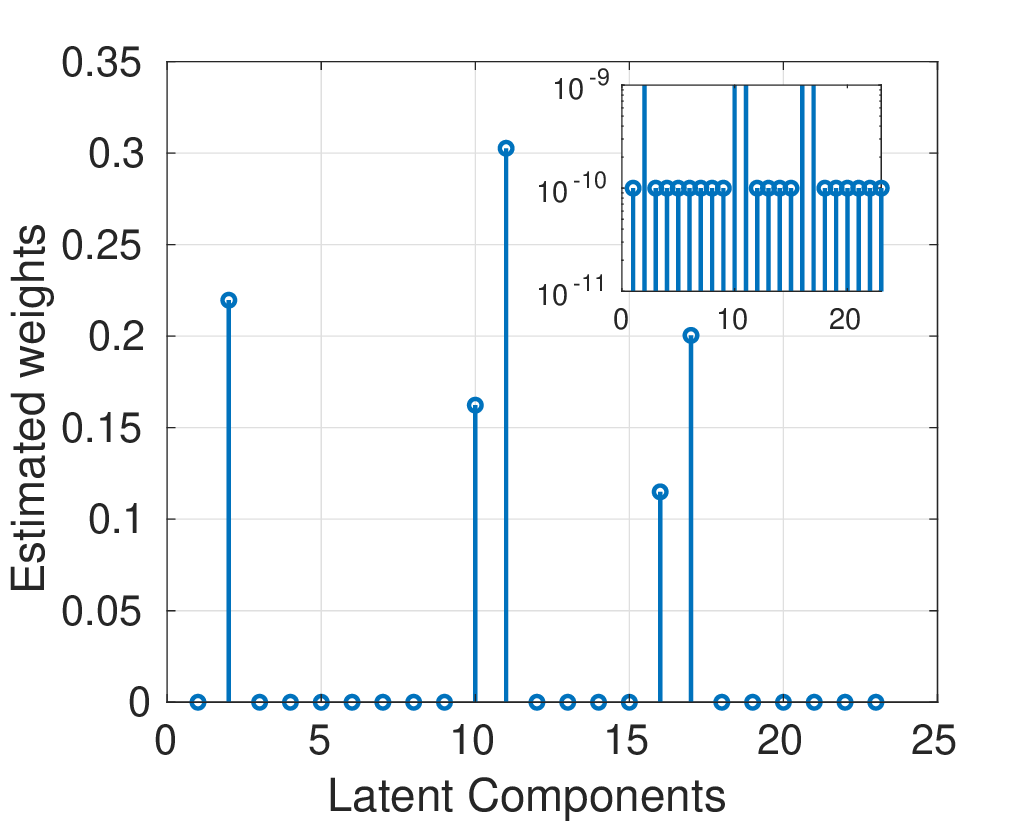}
    \end{tabular}

    \caption{\rev{Typical convergence behavior of the VB-PMF algorithm, initialized with $R=23$ components, for a rank-5 PMF tensor.
    \textbf{Left}: convergence of the evidence lower bound (ELBO) (left y-axis) and evolution of the rank (right y-axis). 
    \textbf{Right}: elements of the estimated loading vector $\widehat{\lambdab}$ after convergence.}}
    
    \label{fig:elbo_conv}
\end{figure}
where we have used \eqref{eq:alpha_lambda} and \eqref{eq:z_est} together with the fact that $\sum_{t,r}\rho_{r,t}=T$. If the $r$th component is not responsible for explaining the observations, then $\sum_{t} \rho_{r,t} \approx 0$. Therefore, for such a component, if $\alpha_{\lambda} \rightarrow 0$, then $\widehat{\lambda}_r \approx 0$ and the entire rank-one component (i.e., $\widehat{\lambda}_r$ and $\{\widehat{\a}_{n,r}\}_{n=1}^N$) can be pruned out of the model. 
\rev{The pruning threshold $\alpha_{\lambda}/T$ is obtained from \eqref{eq:lambda_est_v2} by noticing that when  $\sum_{t} \rho_{r,t} \approx 0$, then $\widehat{\lambda}_r \approx \alpha_{\lambda}/T$ given that $T \gg R\alpha_{\lambda}$.}

The rank can therefore be determined via a single training run \rev{(i.e., without cross-validation)} in which the algorithm is initialized with $R > R_{\mathrm{true}}$ components (where $R_{\mathrm{true}}$ is the unknown true rank of the PMF tensor), followed by \rev{selecting} components with a value \rev{greater than $\alpha_{\lambda}/T$} after convergence. One strategy for choosing the initial number of components is to set $R$ as the maximum possible rank for which the Kruskal uniqueness conditions for the CPD are satisfied \cite{kruskal_three-way_1977}, \cite{sidiropoulos_uniqueness_2000}. Alternatively, the initial rank can be set manually for computational efficiency. 
The algorithm is deemed to have converged when the difference between successive ELBO values is less than a specified threshold. 

\rev{Fig.~\ref{fig:elbo_conv} illustrates the typical convergence behavior of the VB-PMF algorithm. Here, we consider a rank-5 PMF with $N=5$ and $I_n=10, \forall n$. The low-rank components are estimated from a dataset having $T=10^4$ observations and outage probability $p=0$. It can be observed that the ELBO increases monotonically. This corresponds to a successive decrease in the KLD between the estimated and true posterior distributions. On the other hand, the \revFive{inferred} rank decreases from its initial value of 23 components\footnote{For an $N$-dimensional tensor with factor matrices $\A_n \in \RR^{I_n \times R}\revThree{,} \, \forall n$, the Kruskal condition $\sum_n \revFour{k_{\A_n}} \ge 2R + (N-1)$ is sufficient for a unique decomposition \cite{sidiropoulos_uniqueness_2000}. \revFour{Here, $k_{\A_n}$ denotes the Kruskal rank of the matrix $\A_n$, i.e., the largest integer such that every $k_{\A_n}$ columns are linearly independent. Note that if the columns of $\A_n$ are drawn independently from an absolutely continuous distribution, then $k_{\A_n} = \mathrm{min}(I_n,R)$} almost surely. \revFour{Thus,} for the setup considered, the maximum possible rank is $R=23$. \label{foot:max_rank}} to converge to the true value. The stem plot shows the expected value of the loading vector $\lambdab$ after convergence, demonstrating that the desired latent components have been successfully detected while the extraneous components are completely suppressed. Having set $\alpha_{\lambda}=10^{-6}$, the inset plot shows that these extraneous components converge to the value $\alpha_\lambda/T= 10^{-10}$ as expected.}

The complexity of the VB-PMF algorithm consists of updating the parameters of the variational distributions and computing the ELBO to check for convergence. The parameters $\rho_{r,t}$ are computed \revFive{(per iteration)} in $\mathcal{O}(NRT)$ while the combined complexity of computing $\widetilde{\alphab}_{\lambda}$ and $\widetilde{\alphab}_{n,r}$ is $\mathcal{O}(NR)$. In addition, the complexity of computing the ELBO is $\mathcal{O}(NRT)$. Thus, the overall complexity is $\mathcal{O}(NRT)$ \revFive{per iteration}.

\section{\rev{Synthetic-Data Experiments}} \label{sec:synthetic}

In this section, we evaluate the performance of the \rev{proposed PMF estimation approach using} carefully designed synthetic data simulations \rev{to demonstrate its effectiveness in joint parameter and rank inference.}
%
%
We consider a small-scale experiment with $N = 5$ discrete random variables each taking $I_n = 10$ states. We assume that the 5-dimensional joint PMF tensor $\Xt$ admits a na\"{i}ve Bayes model.
The elements of $\lambdab \in \RR ^ R$ and $\{\A_n\}_{n=1}^N \in \RR^{I_n \times R}$ are drawn randomly from $U(0.3,1)$ 
and $U(0,1)$, respectively, followed by normalization according to the probability simplex constraints.
\rev{Sampling the elements of $\lambdab$ from $U(0.3,1)$ ensures that the rank-one component associated with the smallest element in $\lambdab$ contributes significantly to the rank of the PMF tensor.}
The tensor $\Xt$ is then constructed according to \eqref{eq:tensor_element} and is fixed for the rest of the experiment.
For each trial, a dataset $\Y \in \RR^{N\times T}$ containing observations $\y_t$ is obtained by first sampling $\x_t$ from $\Xt$ (according to the na\"{i}ve Bayes model), then randomly and independently zeroing out its elements according to the outage probability $p$ (cf. Section \ref{sec:prel}). 
The accuracy of the estimate is evaluated in terms of the \rev{mean} KLD \rev{and the mean squared relative error (MSRE)} between the true PMF $\Xt$ and the estimated PMF $\widehat{\Xt}$, computed as

\begin{gather*}
    \rev{\mathrm{KLD}_{\mathrm{mean}}} = \rev{\revFive{\hat{\mathbb{E}}}}\left[\sum_{i_1,\dotsc,i_N} \Xt(i_1,\dotsc,i_N) \log \frac{\Xt(i_1,\dotsc,i_N)}{\widehat{\Xt}(i_1,\dotsc,i_N)}\right] \\
    \rev{\mathrm{MSRE} = \revFive{\hat{\mathbb{E}}}\left[\frac{\|\Xt-\widehat{\Xt}\|_\mathsf{F}^2}{\|\Xt\|_\mathsf{F}^2}\right],}
\end{gather*}
\revFive{where $\hat{\mathbb{E}}$ denotes the empirical average over all trials.}

We initialize VB-PMF with 23 components (see footnote~\ref{foot:max_rank})  and set the hyperparameters $\alpha_{\lambda}$ and $\alpha_{n,r}$ to $10^{-6}$ and $1$, respectively. This hyperparameter setting favors sparse solutions for $\lambdab$ and all possible solutions for $\{\A_n\}_{n=1}^N$, allowing the rank to be estimated as part of the inference.

The performance of VB-PMF is compared with \rev{various PMF estimation algorithms and model order selection techniques as described below:}
\rev{\begin{itemize}
    \item \textbf{SQ-AIC, SQ-BIC, SQ-DNML:} The PMF is estimated using SQUAREM-PMF \cite{chege_efficient_2022} which is an accelerated form of the expectation maximization (EM) algorithm \cite{yeredor_maximum_2019}. The best rank is selected from a list of candidates $\{1,\dotsc,R+5\}$ using the AIC \cite{akaike1974new}, the BIC \cite{schwarz1978estimating}, and the DNML \cite{yamanishi_decomposed_2019} likelihood-based criteria. 
    \item \textbf{CTF3D-ValErr, CTF\revOneB{-Full}-ValErr:} The PMF is estimated using the coupled tensor factorization (CTF) approach with a KLD loss criterion \cite{kargas_tensors_2018}, \cite{pmlr-v89-kargas19a}. We consider joint PMF recovery based on \revOneB{empirical} third-order marginals (CTF3D) as well as the full \revOneB{emprical} PMF (CTF\revOneB{-Full}). The best rank is selected from a list of candidates $\{1,\dotsc,R+5\}$ by observing validation errors (ValErr) on a small held-out dataset. 
    \item \textbf{SQ-Thresh (Low), SQ-Thresh (High):} The PMF is estimated using SQUAREM-PMF \cite{chege_efficient_2022} initialized with 23 components as with VB-PMF. For SQ-Thresh (low), we remove rank-one components \revFive{whose estimated weight (probability) is} less than 10\% of the largest element in the estimated loading vector. \revFive{Therefore, this approach serves as a na\"{i}ve baseline for model order selection.} On the other hand, for SQ-Thresh (High), we assume that the true rank $R$ is known and keep only the largest $R$ components after estimation. In both cases, we renormalize the CPD components to restore the probability simplex constraints. 
    \item \revOneB{\textbf{TN-ValErr:} The PMF is estimated using a tensor network (TN) approach \cite{glasser2019expressive}. Specifically, we use the locally purified state (LPS) representation, which the authors recommend to be preferred over other TN representations (such as tensor trains (matrix product states) and Born machines) in general. Following \cite{glasser2019expressive}, we choose a purification dimension of 2. The best LPS purification rank is selected from the range $\{2,4,6,8,10\}$ by observing validation errors. Note that since the interpretation of the purification rank is different from that of the CPD rank, we only compare the MSRE, KLD, and runtime performance.}
\end{itemize}}

\rev{We present results averaged over 50 independent trials.}

\subsection{\rev{Performance with varying number of observations}} \label{subsec:results_A}
\begin{figure*}[t]
    \centering

    \begin{tabular}{cc}
        \includegraphics[width=0.38\textwidth]{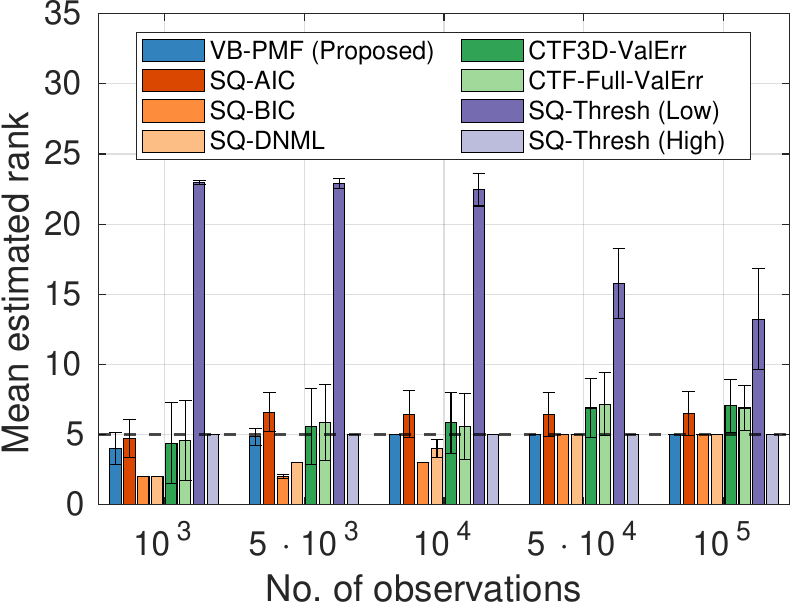} &
        \includegraphics[width=0.4\textwidth]{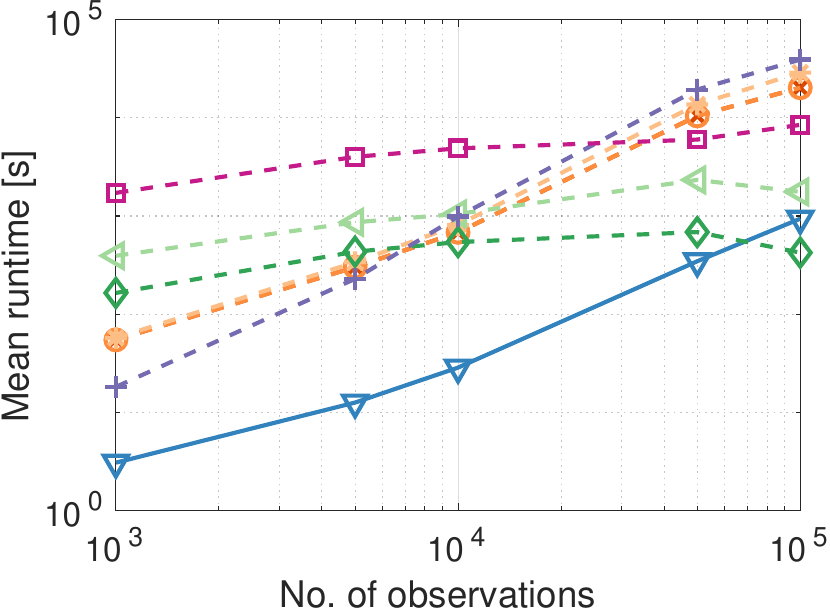} \\[1em]
        \multicolumn{2}{c}{\includegraphics[width=0.95\textwidth]{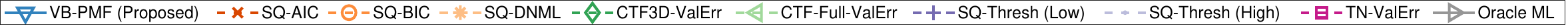}} \\[1em]
        \includegraphics[width=0.4\textwidth]{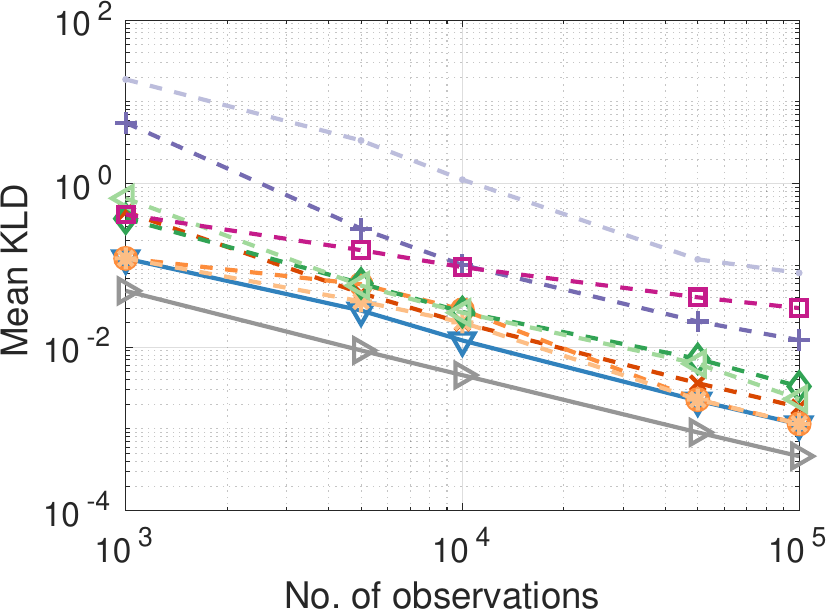} &
        \includegraphics[width=0.4\textwidth]{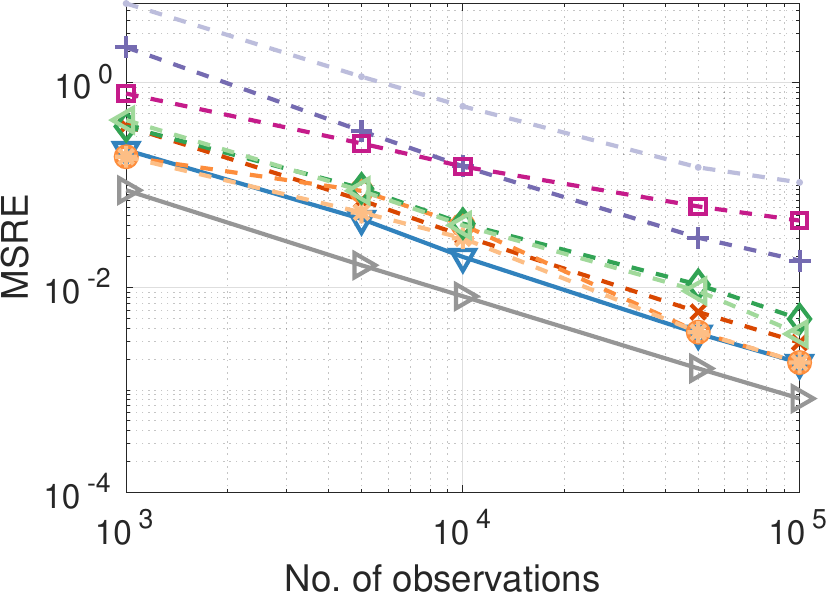}
    \end{tabular}

    \caption{Joint PMF estimation performance of VB-PMF, compared to other PMF estimation approaches, versus the number of observations $T$, averaged over 50 trials.}
    \label{fig:exp_1}
\end{figure*}

Fig. \rev{\ref{fig:exp_1}} presents results for a rank-$5$ PMF tensor \rev{and an outage probability $p = 0$, i.e., no missing observations.} \rev{We observe that} the rank inferred by VB-PMF converges to the true rank as the number of observations is increased. \rev{This demonstrates empirically that VB-PMF is statistically consistent\footnote{A consistent model selection criterion is one which almost surely (i.e., with probability one) selects the candidate model having the correct dimension as the sample size grows infinitely, given that the dimension of the true model is represented among the candidates \cite{cavanaugh_akaike_2019}.}. We emphasize that the rank is estimated from the data as part of the inference procedure without cross-validation.}
\rev{Both} \rev{SQ-}BIC and \rev{SQ-}DNML \rev{are consistent criteria and therefore detect the true rank as $T$ increases}. On the other hand, \rev{SQ-}AIC \rev{and both CTF approaches exhibit a large variance in the inferred rank. In addition, thresholding as applied in SQ-Thresh (Low) is not able to estimate the correct rank, although the error decreases as $T$ grows.}

\rev{From the MSRE and mean KLD metrics}, VB-PMF performs comparably to or better than the other model order estimation techniques. Here, the `Oracle' ML estimate, which assumes that the true hidden state associated with each observation is known, serves as an empirical lower bound for the \rev{MSRE and the} KLD. This result can be attributed to the fact that VB-PMF detects the correct rank with minimal or no misfit as the size of the dataset increases.
\rev{The underfitting observed in VB-PMF for smaller values of $T$ occurs due to the Bayesian algorithm preferring simpler models (i.e., with fewer parameters) in small-sample regimes and instead seeking a closer fit to the data. Such accuracy-complexity tradeoffs are a key part of the AIC, BIC, and DNML criteria; however, with the Bayesian approach, they occur implicitly during the inference procedure, making manual model order selection unnecessary.}
\rev{SQ-Thresh (High) has the lowest accuracy despite \revFive{using} the \revFive{true} model order (which, for illustration purposes, was assumed to be known in advance) due to a poor fit to the data.} 
Moreover, the runtime plots show that VB-PMF provides significant gains in computational efficiency, even though the runtime increases linearly with the dataset size \revFour{$T$}. This is because the rank is estimated via a single training run, while the other \rev{algorithms} need \rev{to be trained afresh for} each candidate rank. \rev{The computational complexity of the CTF approaches is dominated by the \revFive{computation} of \revFive{the joint CPD from the} \revOneB{empirical estimates}. Hence, it does not grow as rapidly with $T$ as VB-PMF. However, for larger dataset sizes, VB-PMF still exhibits an advantage over CTF in terms of accuracy.} \revOneB{Although TN representations such as the LPS model possess strong expressive power, VB-PMF achieves higher accuracy since the generated data arise from a low-rank structure that closely matches the assumptions underlying our method. In addition, selecting the TN rank via cross-validation increases the computational cost in terms of the runtime.}


\rev{\subsection{Performance with varying ranks and outage probabilities}} \label{subsec:results_B}
\begin{figure*}[t]
    \centering
    \includegraphics[width=0.9\textwidth]{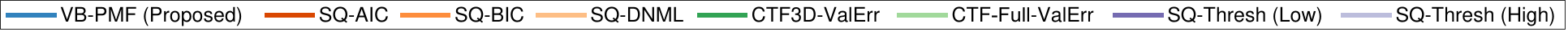}
    \par \vspace{0.8em}
    \includegraphics[width=0.31\textwidth]{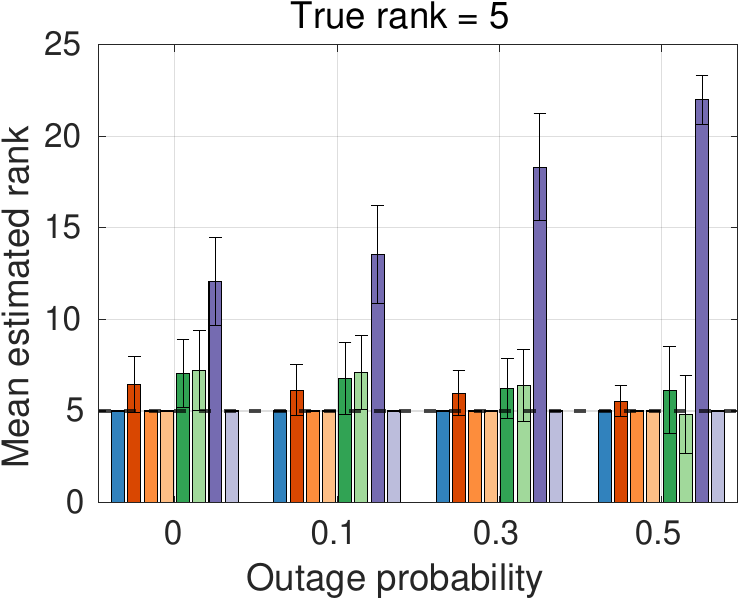}
    \hfill
    \includegraphics[width=0.31\textwidth]{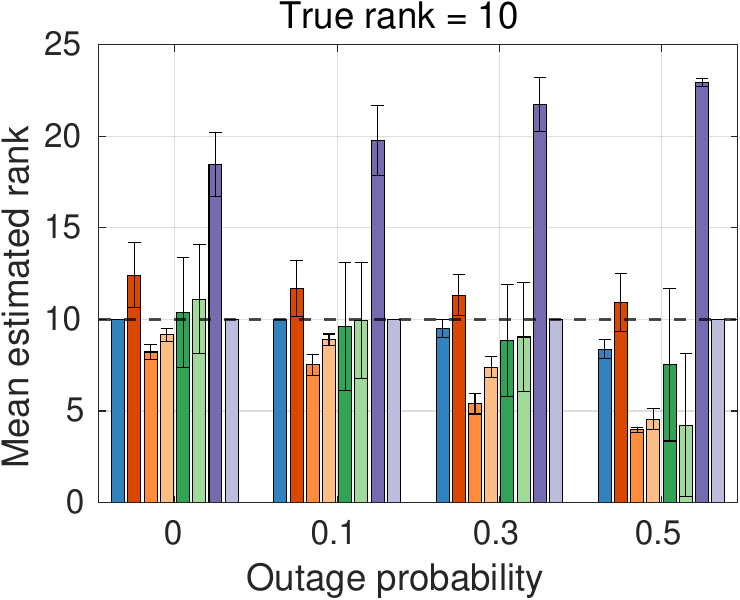}
    \hfill
    \includegraphics[width=0.31\textwidth]{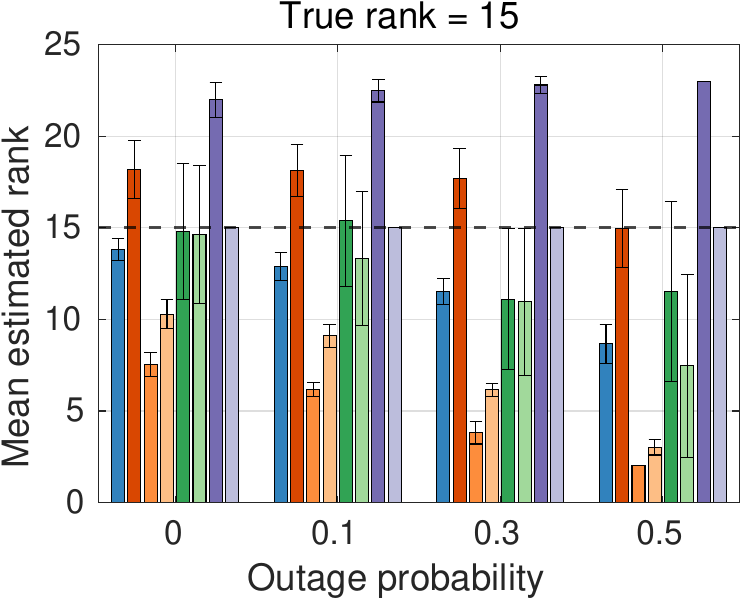}

    \caption{Rank estimation performance versus the outage probability $p$.}
    \label{fig:exp_2a}
\end{figure*}

\begin{figure*}[t]
    \centering
    \hspace{2em}
    \includegraphics[width=0.95\textwidth]{figures/lgnd_line.pdf}
    \par \vspace{0.8em}
    \includegraphics[width=0.31\textwidth]{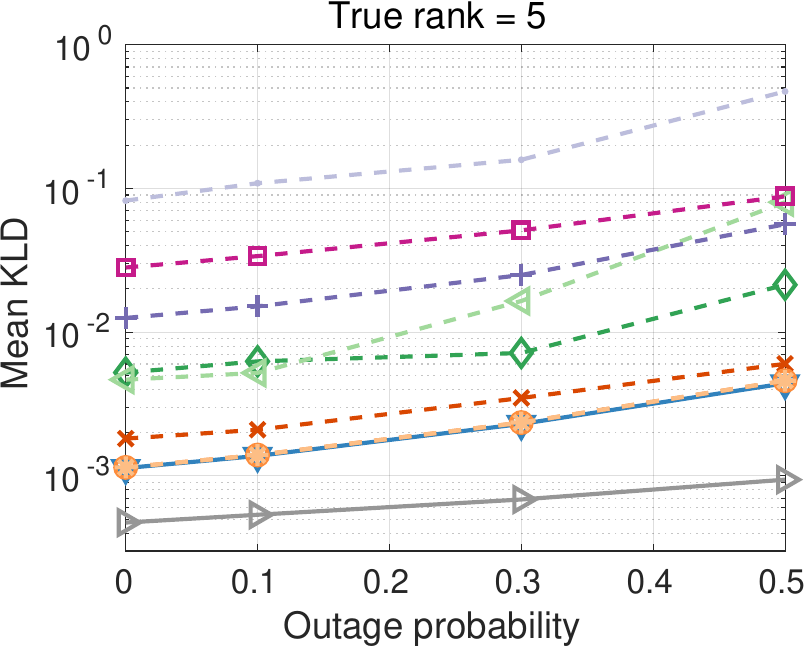}
    \hfill
    \includegraphics[width=0.31\textwidth]{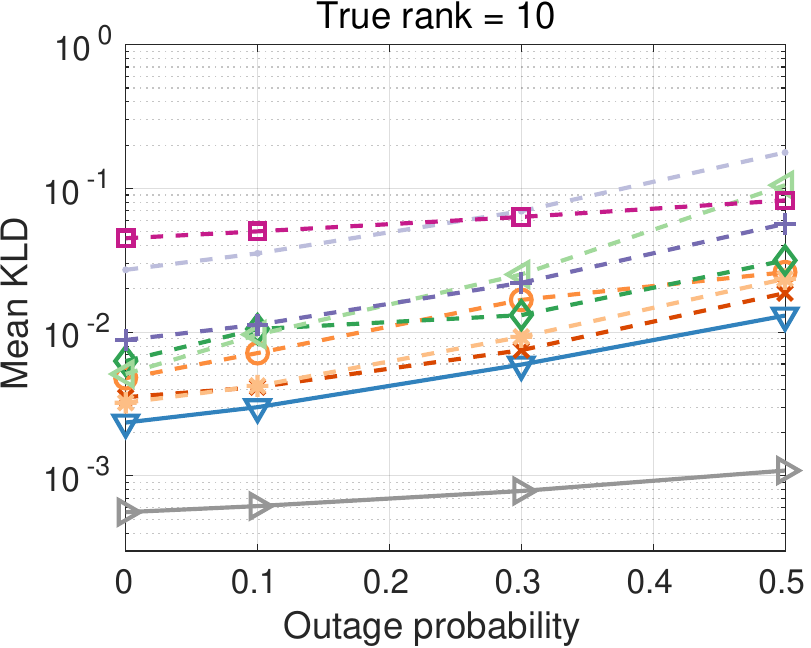}
    \hfill
    \includegraphics[width=0.31\textwidth]{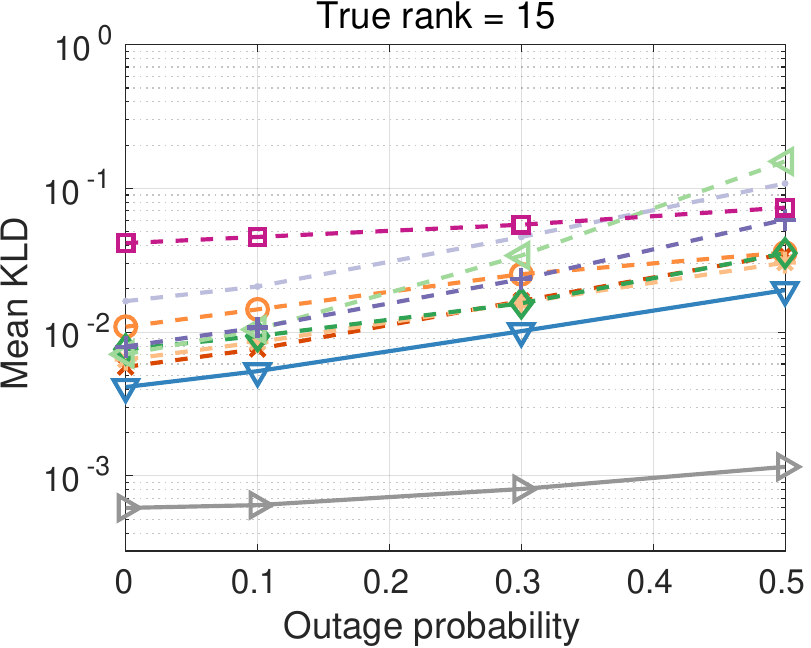}
    \caption{Mean KLD between the estimated and true PMF tensors versus the outage probability $p$.}
    \label{fig:exp_2b}   
\end{figure*}
\rev{Next, we study the performance of VB-PMF for various ranks and outage probabilities. 
The ranks considered are $R = \{5,10,15\}$, the outage probabilities are $p=\{0,0.1,0.3,0.5\}$, while the number of observations is set to $T=10^5$.
Figs.~\ref{fig:exp_2a} and \ref{fig:exp_2b} presents the mean estimated rank and the mean KLD as a function of the outage probability. The MSRE plots showed a similar trend to the KLD plots and were therefore not included for brevity. For $R=5$, VB-PMF estimates the correct rank for all outage values. This also holds for $R=10$ and $p=\{0,0.1\}$. In these cases, even with missing observations, the data is sufficient to enable the algorithm to correctly learn the low-rank structure. In the remaining cases, the underfitting by VB-PMF occurs due to the accuracy-complexity tradeoff described in Section~\ref{subsec:results_A}. This can be understood more clearly in Fig.~\ref{fig:exp_2b}, where VB-PMF achieves the best KLD performance in all cases except $R=5$, where SQ-BIC and SQ-DNML also show similar results. Note that SQ-BIC and SQ-DNML estimate the correct rank across all outage values for $R=5$. However, for higher rank and outage values, these two techniques prefer smaller models and heavily penalize models with more parameters, which affects their accuracy. In comparison, SQ-AIC selects larger models due to the weaker AIC penalty. A large variance in the estimated rank is observed when using validation errors as in the CTF approaches, especially for higher ranks and outage probabilities. Moreover, as the outage increases, the accuracy of the CTF methods is affected due to less reliable \revOneB{empirical} PMF estimates, particularly for CTF\revOneB{-Full}-ValErr.
\revOneB{Although VB-PMF is more accurate than TN-ValErr across the considered outage probabilities, the performance degradation of TN-ValErr appears comparatively smaller as the fraction of missing observations increases. This suggests that TN models may exhibit greater robustness at high outage probabilities, possibly due to their high expressive power.}
Overall, we conclude that VB-PMF provides the best accuracy-complexity tradeoff while also estimating the rank without cross-validation. 
\begin{figure}[t]

    \begin{tabular}{cc}
        \hspace{-1em}
        \includegraphics[width=0.48\columnwidth]{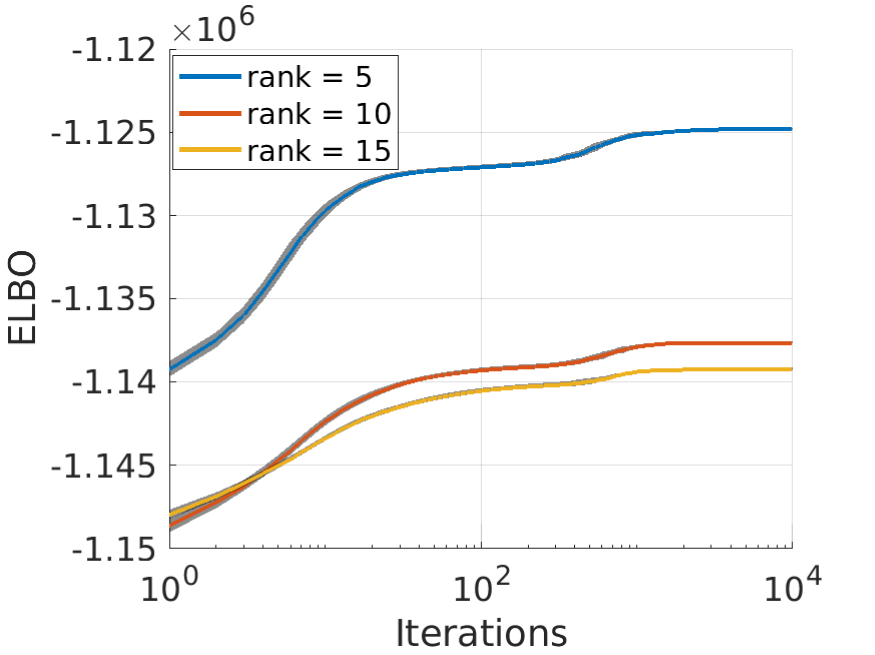} &
        
        \hspace{-1em}
        \includegraphics[width=0.45\columnwidth]{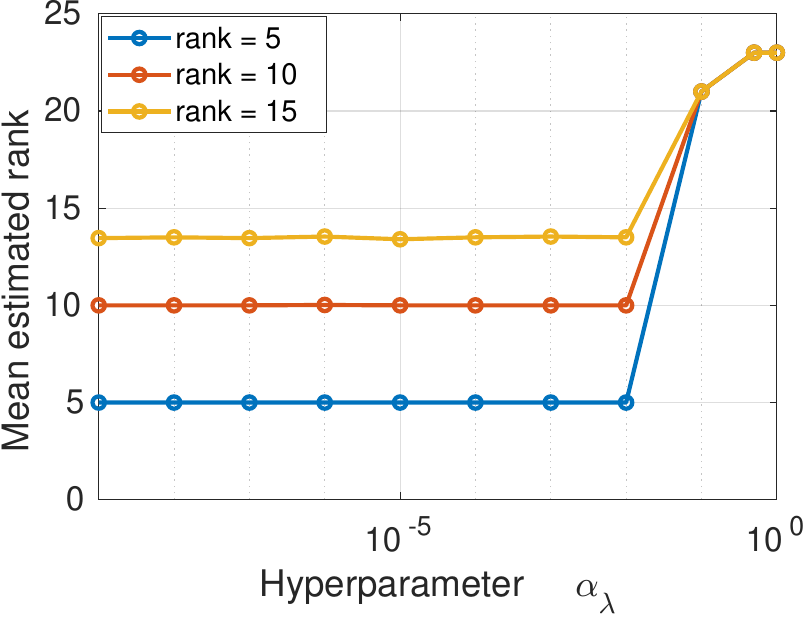}
    \end{tabular}

    \caption{\rev{
    Sensitivity of VB-PMF to the hyperparameter $\alpha_{\lambda}$ for $T=10^5$ and $p=0$. 
    \textbf{Left:} The evidence lower bound (ELBO) averaged over various values of $\alpha_{\lambda}$. 
    The shaded region represents one standard deviation away from the mean. 
    \textbf{Right:} Mean estimated rank versus different values of $\alpha_{\lambda}$.
    }}
    
    \label{fig:sensitivity}
\end{figure}
\begin{figure}
\centering
\includegraphics[width=0.95\columnwidth]{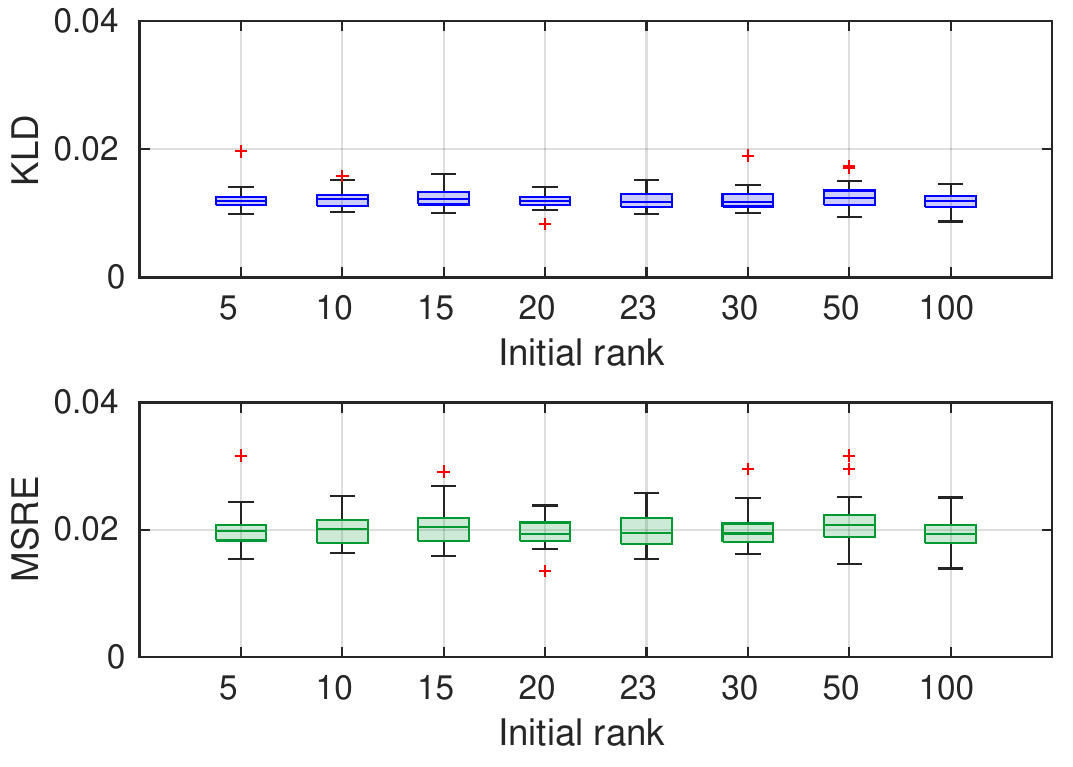}
    \caption{\revOneB{Effect of different values of the initial VB-PMF rank on the accuracy (in terms of the KLD and the MSRE) of the estimated joint PMF.}}
    \label{fig:init_rank_sensitivity}
\end{figure}

\revOneB{\subsection{Effect of the choice of the hyperparameter $\alpha_\lambda$ and the initial VB-PMF rank}}
In Fig.~\ref{fig:sensitivity}, we investigate the effect of the choice of the hyperparameter $\alpha_{\lambda}$ on the ELBO convergence and the rank estimation performance of VB-PMF. The values considered are $\alpha_{\lambda}=\{10^{-9}, 10^{-8},\dotsc,10^{-1}, 0.5,1\}$ for $T=10^5$ observations and outage probability $p=0$. The ELBO figure shows the mean, taken over the values of $\alpha_{\lambda}$, and the region of one standard deviation from the mean. We observe that the convergence of the ELBO does not depend significantly on the choice of $\alpha_{\lambda}$. In addition, the estimated rank is constant for $\alpha_{\lambda} \le 10^{-2}$ and is accurately obtained for $R=\{5,10\}$. For $R = 15$, some underfitting occurs due to insufficient data, as already seen in Fig.~\ref{fig:exp_2a}. However, setting $\alpha_{\lambda} = \{10^{-1}, 0.5, 1\}$ results in less sparse loading vector estimates, leading to rank estimates that are higher than required. A closer inspection of the estimated loading vectors in these cases revealed that although the unnecessary latent components were much smaller than the desired ones, they were not sufficiently suppressed, resulting in values greater than the pruning threshold $\alpha_{\lambda}/T$. Therefore, these components could not be removed without manual inspection or an \revFive{\textit{ad hoc}} specification of the pruning threshold. In conclusion, it is recommended to set $\alpha_{\lambda} \ll 1$ to enable ``automatic" pruning of unnecessary latent components.}

\revOneB{We also investigate the impact of the initial rank on the KLD and the MSRE performance. Fig.~\ref{fig:init_rank_sensitivity} displays the results for $T=10^4$, $p=0$, and $R=5$, averaged over 50 trials. We observe that the initial rank has a negligible effect on the estimation accuracy. In this experiment, we found that VB-PMF recovered the correct rank in virtually all cases. This result demonstrates that VB-PMF is robust to the choice of the initial rank in scenarios where the low-rank assumption is fulfilled, as in this controlled experiment.}

\section{Real-Data Experiments} \label{sec:real}

\rev{In this section, we illustrate the effectiveness of VB-PMF for various statistical learning tasks using real data. Specifically, we evaluate performance in data classification and item recommendation (rating prediction) tasks. For comparison, \revOneB{unless otherwise stated,} we consider SQ-AIC, SQ-BIC, SQ-DNML, CTF3D-ValErr\revOneB{, CTF-Full-ValErr}, \revOneB{TN-ValErr,} (see Section~\ref{sec:synthetic} for a description), as well as some commonly used classical approaches which serve as \revFive{benchmarks} for each task.
\subsection{Data Classification} \label{subsec:classification}
For this task, we select 5 datasets from the UCI Machine Learning Repository \cite{UCIRepository}, namely, Credit Approval (\textsc{Credit}), Occupancy Detection (\textsc{Occupancy}), Congressional Voting Records (\textsc{Votes}), \textsc{Iris}, and Website Phishing (\textsc{Website}). The first three datasets contain 2 classes (binary classification) while the last two datasets have 3 classes (multiclass classification). In order to apply our approach (and other PMF estimation algorithms), any continuous features are discretized. \revOneB{Table~\ref{tab:summary_UCI} summarizes the datasets used in the classification experiments.}

Our goal is to estimate the class label of test samples using the joint PMF of the features and the labels, estimated from training data. Given an estimated joint PMF model $\{\widehat{\lambdab},\{\widehat{\A}_n\}_{n=1}^N\}$ and a test sample $\y_{\rm test} = [y_1,\dotsc,y_{N-1}, y_N]$ where $y_N$ is the unknown class label, an estimate is found via MAP estimation
\begin{table}[] 
\normalsize
\caption{\revOneB{Summary of the datasets used in the classification experiments. The number of random variables is denoted by $N$ (i.e., $N-1$ features and one label) while $T$ is the number of instances (observations).}}
\label{tab:summary_UCI}
\centering
\begin{tabular}{l|c|c|c}
Dataset   & $N$ & $T$ & Classes \\ \hline
\textsc{Credit}    & 10              & 690               & 2       \\
\textsc{Occupancy} & 6               & 20560             & 2       \\
\textsc{Votes}     & 17              & 435               & 2       \\
\textsc{Iris}      & 5               & 150               & 3       \\
\textsc{Website}   & 10              & 1353              & 3    
\end{tabular}
\end{table}
\begin{equation}
\begin{aligned}
    \widehat{y}_N & = \argmax{i_N \in \{1,\dotsc,I_N\}} \mathsf{Pr}(i_N\,|\,y_1,\dotsc,y_{N-1}) \\
    & = \argmax{i_N \in \{1,\dotsc,I_N\}} \sum_{r=1}^R \widehat{\lambda}_r \widehat{\A}_N(i_N,r)\prod_{n=1}^{N-1} \widehat{\A}_n(y_n,r).
\end{aligned}
\end{equation}

For VB-PMF, an 80\%\,/\,20\% train\,/\,test split is applied to each dataset. \revOneB{The initial ranks for \textsc{Credit}, \textsc{Occupancy}, \textsc{Votes}, and \textsc{Iris} are set to 16, 98, 17, and 19, respectively. The initial ranks correspond to the maximum Kruskal rank for each dataset's low-rank PMF model. A similar initialization for \textsc{Website} ($R = 10$) resulted in the same estimated rank after running VB-PMF, possibly indicating a true rank larger than the maximum Kruskal rank. Therefore, for \textsc{Website}, we set the initial rank to $R=50$.} \revFive{For} the other PMF estimation algorithms, \revFive{a 70\%\,/\,10\%\,/\,20\% train/validate/test split is applied, where 10\% of the data forms} a validation set for rank estimation. \revOneB{We test candidate ranks in the range $\{1,2,\dotsc,20\}$.} The results are averaged over 50 independent trials, each with a different dataset split. As a \revFive{benchmark}, we consider the well-established random forest (RF) classifier \cite{breiman2001random} which is known to perform \revFive{particularly} well on tabular data. The predictive performance of the algorithms is evaluated in terms of the classification accuracy and the F1~score \cite{sokolova2009systematic}, which measures how a classification model balances \revFive{the correct identification of} positive instances (precision) with \revFive{the minimization of} false positives and false negatives (recall)\footnote{\revFive{For the multiclass classification experiments, we compute the F1 score based on the per-class average precision and recall. This is one of the variants of the so-called macro-F1 score \cite{sokolova2009systematic}.}}.
\begin{figure}
    \centering
    \includegraphics[width=0.95\columnwidth]{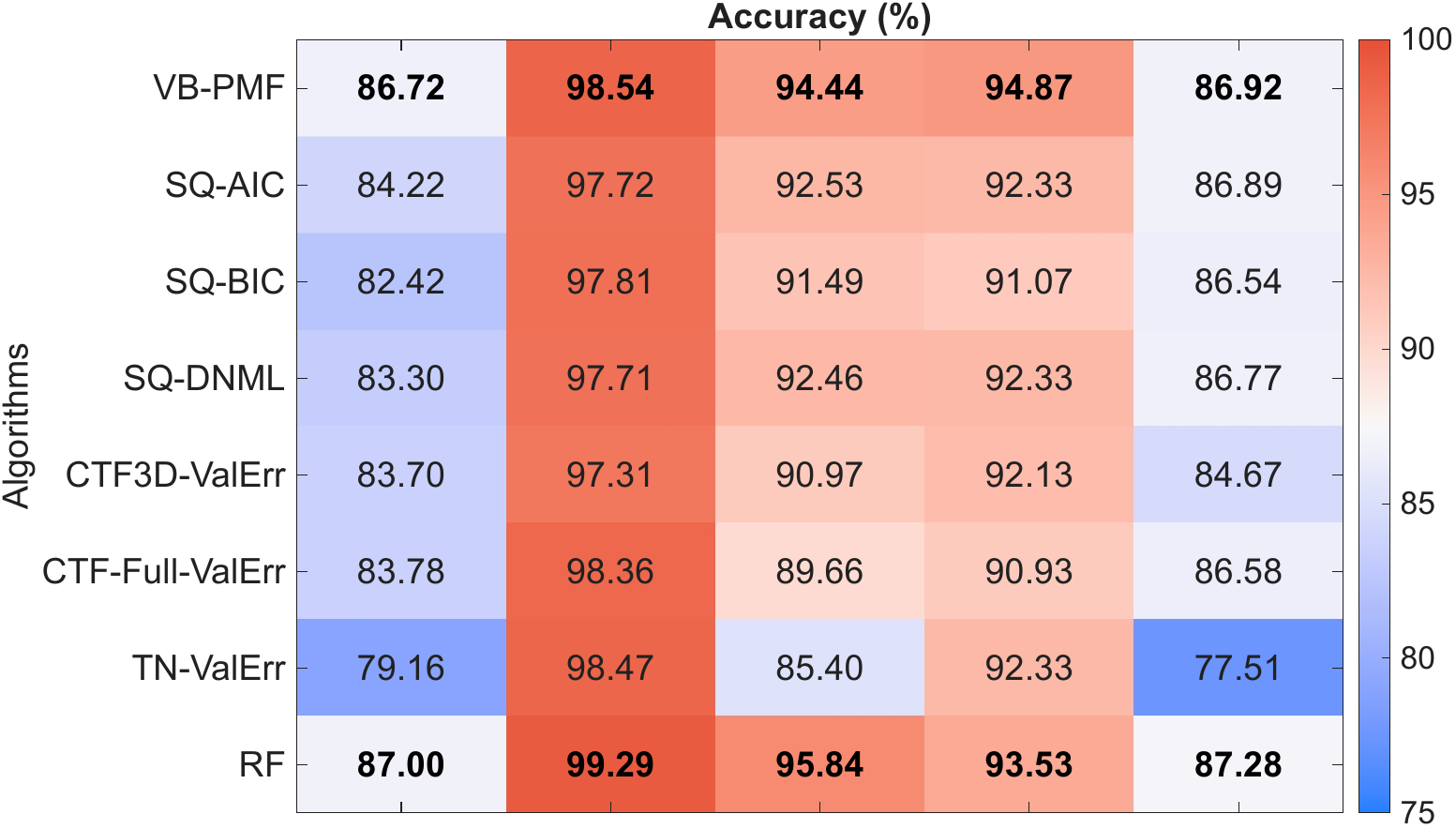} \\
    \vspace{1em}
    \includegraphics[width=0.95\columnwidth]{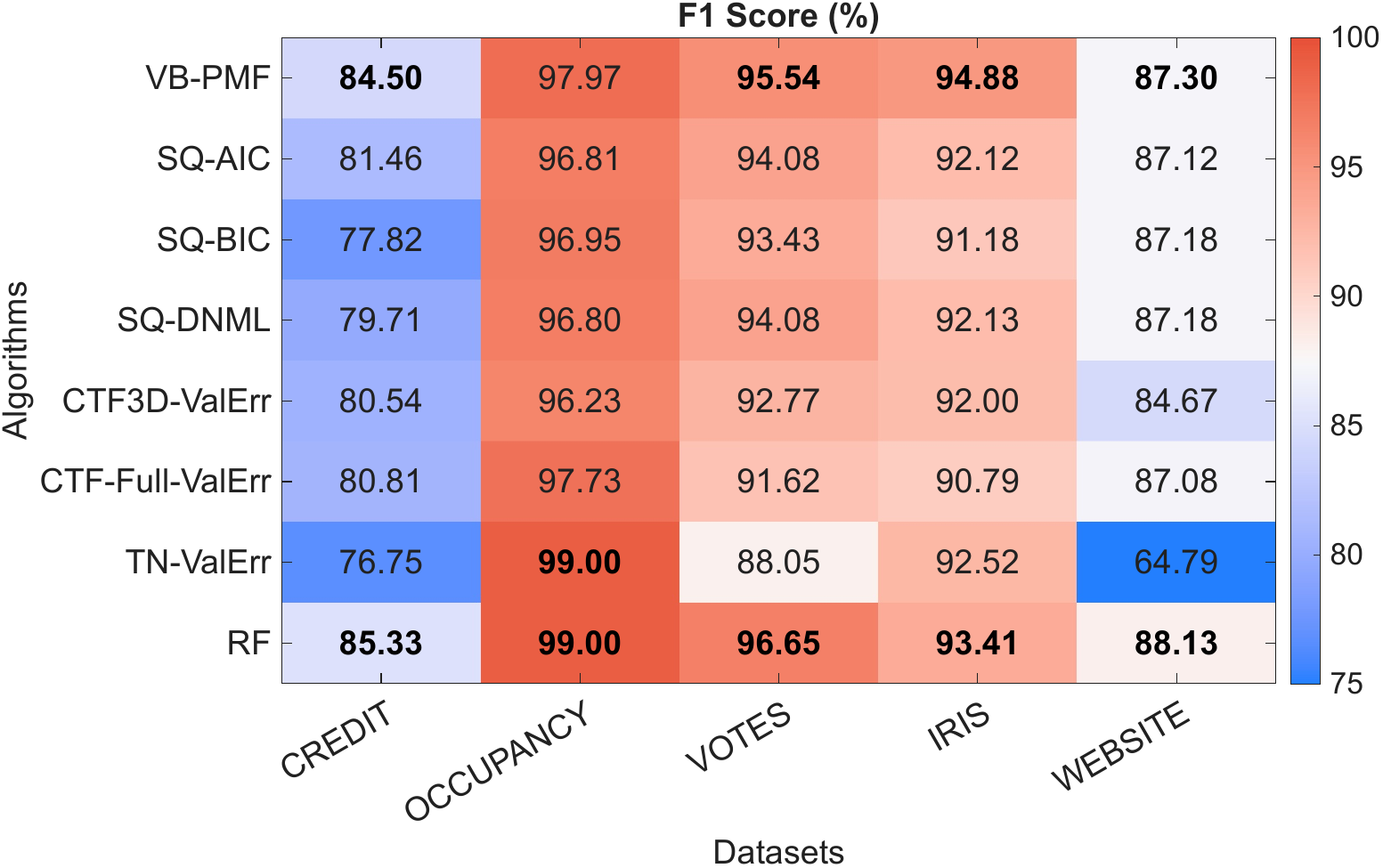}
\caption{\rev{Classification accuracy and F1~score, averaged over 50 trials, on various datasets from the UCI Machine Learning Repository. Results from the top two algorithms are highlighted in bold. Our proposed method (VB-PMF) achieves results comparable to the random forest \revFive{benchmark} across all datasets.}}
\label{fig:classification_results}
\end{figure}

Fig.~\ref{fig:classification_results} shows the classification accuracy and the F1~score achieved by all approaches on the UCI datasets. Firstly, \revOneB{in terms of accuracy,} VB-PMF outperforms the other PMF-based approaches across all datasets. \revOneB{VB-PMF also achieves the best F1 score (among PMF-based approaches) in four out of five datasets.} The Bayesian model order selection mechanism integrates over parameter uncertainty rather than relying on point estimates of the parameters, leading to more robust and stable model selection. Moreover, the Bayesian formulation automatically balances model fit and complexity through the marginal likelihood, thus avoiding the need for \revFive{\it{ad hoc}} model selection procedures such as cross-validation or information criteria like AIC, BIC, and DNML. \revOneB{Notably, TN-ValErr exhibits a large variance in performance under the parameter settings we consider (we use the same settings for all datasets). This suggests that careful tuning of the LPS model parameters (e.g., purification dimension, purification rank, batch size, learning rate, etc.) is necessary to achieve consistent performance. VB-PMF does not require such parameter tuning, and the ``default" hyperparameter settings used in the synthetic-data experiments ($\alpha_{n,r}=1$, $\alpha_\lambda=10^{-6}$) work quite well here.} 

Beyond comparisons within the PMF family, it is also useful to evaluate VB-PMF in relation to a widely used classifier such as the random forest (RF). VB-PMF \revOneB{outperforms RF on the \textsc{Iris} dataset and} achieves classification accuracies and F1 scores within approximately 1\% of the RF \revFive{benchmark} across \revOneB{four} datasets, even though it is a general-purpose probabilistic model not specifically designed for classification. These results indicate that the joint PMF model learned by VB-PMF is effective for classification tasks. In addition to its competitive predictive performance, VB-PMF offers advantages that purely discriminative models like RF do not provide. By estimating the full joint PMF of the features and labels, VB-PMF yields a richer representation of the data than discriminative models that approximate only conditional class probabilities. 

Fig.~\ref{fig:example_insights} exemplifies some insights from the Congressional Voting Records dataset \cite{UCIRepository} using a low-rank model estimated by VB-PMF. The probability of belonging to a certain party (class) given the latent components reveals various voting blocs such as ``strong Republican" (components 1, 3, and 6), ``strong Democrat" (components 2 and 4), ``bipartisan" (component 5), and ``slightly Republican" (component 7). Furthermore, considering the probability of voting ``Yes" on a certain issue (feature) given the latent components, some meaningful ideological differences within each party emerge. For instance, although blocs 1 and 3 are Republican, they differ on issues such as arms control (anti-satellite-test-ban) and foreign policy (aid-to-nicaraguan-contras). Similarly, there are clear differences between Democrat blocs 2 and 4 on crime and foreign policy. Finally, Republican bloc 6 consists of moderates who mostly take centrist positions. Such insights are not available to purely discriminative models such as RF, highlighting the added benefit of low-rank joint PMF models.}
\subsection{Item Recommendation} \label{subsec:recommendation}

\begin{figure}
    \centering
    \includegraphics[width=0.95\columnwidth]{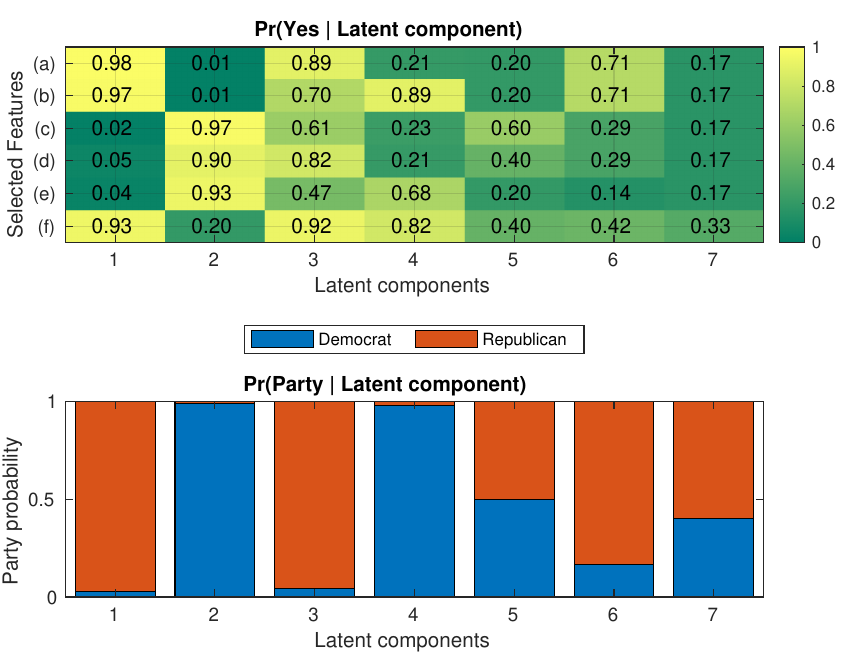}
    \caption{\rev{Insights from the Congressional Voting Records dataset \cite{UCIRepository} using a low-rank model estimated by VB-PMF. The selected features are (a) physician-fee-freeze, (b) el-salvador-aid, (c) aid-to-nicaraguan-contras, (d) anti-satellite-test-ban, (e) adoption-of-the-budget-resolution, and (f) crime. The latent components uncover various voting blocs within the Democrat and Republican classes that purely discriminative models cannot reveal.}}
    \label{fig:example_insights}
\end{figure}
We evaluate the performance of the proposed algorithms in a real-world movie recommendation application. For this purpose, we utilize the MovieLens \rev{10}M dataset \cite{harper_movielens_2016} which contains movie ratings on a 5-star scale with half-star increments, i.e., $\{0.5,\dotsc,5.0\}$. From the entire dataset, we select the ratings of \rev{$N=100$} top-rated movies\footnote{\rev{It is common in collaborative filtering applications to select popular items \revFour{in} a preprocessing step (see, e.g., \cite{aharon2015excuseme, rahdari2025under}). Since we estimate a multivariate PMF of all movie ratings, selecting popular items can be viewed as a dimensionality reduction step. In addition, the accuracy of the estimated PMF depends on the number of data samples (i.e., users) available. Thus, it would not be feasible to model the joint PMF of all 10,000 movies with \revFour{a} relatively small number of users (72,000) in the dataset. Our choice to use the 100 most rated movies results in a movie-user rating matrix with about $94\%$ of the total number of users, and with reasonable dimensions such that a reliable joint PMF estimate can be obtained.}}. 
We perform some preprocessing by omitting users who have rated only one or none of the selected movies and mapping the ratings to a discrete scale $\{1,\dotsc,10\}$. Thus, we \rev{extract} a movie-user rating matrix $\Y \in \RR^{100\times T}$, where the number of users \rev{$T=67,\!741$}.
\revThree{The percentage of missing ratings in the rating matrix is about \rev{$72\%$} (outage probability \rev{$p=0.72$}).}
\revFive{S}ince we are dealing with real data, the true rank (model order) \rev{of the joint PMF} is unknown.

We aim to estimate the joint PMF (that is, $\{\A_n\}_{n=1}^N$ and $\lambdab$) of the ratings \rev{ (i.e., the random variables \revFive{of all 100 movies}) and use the estimated PMF} to predict the missing ratings. In this scenario, the joint PMF is a \rev{100}-dimensional tensor $\Xt \in \RR ^ {I_1\times \cdots \times I_N}$ with each dimension $I_n=10$, $n=1,\dotsc,100$. We run \rev{50} Monte Carlo trials in which 80\% of the ratings are used to estimate the joint PMF, \rev{while} 20\% are used for testing. In each trial, a different training/testing split is applied to the rating matrix.

\revOneB{The maximum Kruskal rank in this experiment is $R=450$. Using this value as an initial rank results in a low-rank model having a large number of parameters, which is computationally expensive to estimate\footnote{\revOneB{To check this assertion, we conducted the experiment with $R=450$ as the initial rank. We obtained somewhat degraded RMSE and MAE values (0.879 and 0.676, respectively) with a significantly higher computational cost.}}. Thus, we choose $R=50$ as the initial VB-PMF rank. For the other PMF estimation approaches, we tested candidate ranks within $\{5,10,15,\dotsc,50\}$ during the validation stage.}

During the testing phase, for each test sample, one rating is hidden and predicted using the estimated PMF. The predicted rating is found by computing its conditional expectation with respect to the estimated joint PMF. In particular, let the current test sample be $\y_{\rm test} = [y_{1}, \dotsc, y_{N}]^\mathsf{T}$ and suppose that $y_{N} = 0$, i.e., hidden. The predicted rating is then

\begin{equation} \label{eq:conditional_exp}
    \widehat{y}_{N} = \sum_{i_N=1}^{I_N} i_N \mathsf{Pr}(i_N\,|\,y_{1}, \dotsc, y_{N-1}),
\end{equation}
where the conditional PMF can be computed from the joint PMF via Bayes' theorem.

\rev{Recall that CTF3D is based on \revFour{a} coupled factorization of third-order marginal PMFs \cite{kargas_tensors_2018} (cf. Section \ref{sec:synthetic}). For this experiment, with $N=100$, computing all $N \choose 3$ third-order marginals is prohibitive in terms of computational complexity and memory requirements. Therefore, we consider a randomly chosen subset consisting of a quarter of all possible triplet combinations. It has been shown in \cite{flores_coupled_2022} that this strategy reduces computational complexity with a relatively minor loss in accuracy.} 

\rev{As a \revFive{benchmark}, we consider biased matrix factorization (BMF) \cite{koren_matrix_2009}, a standard and widely used method for recommend\revFive{er} systems. For BMF, the matrix rank is chosen by observing validation errors (ValErr) using 10\% of the ratings as a validation set. Finally, we also calculate the user average, the movie average, and the global average, which serve as na\"{i}ve predictors of the missing ratings.} The quality of the predictions is evaluated by computing the RMSE and the mean absolute error (MAE) between the predicted and true ratings. 

Table~\ref{table:1} presents \rev{results from the item recommendation experiment}. 
%
\rev{VB-PMF and CTF3D-ValErr achieve comparable results, with the latter having \revFour{a} slightly better MAE performance. However, VB-PMF has a much lower execution time due to the efficient VI-based implementation. In addition, no lower-order marginals need to be computed for VB-PMF and the SQ-based approaches, which also saves computation time. Note that for a fairer \revFour{runtime} comparison, the candidate ranks for CTF3D and the SQ-based methods \revFour{were tested simultaneously by employing parallel computation}. These results further demonstrate the advantages of the Bayesian framework underpinning our approach in terms of providing a good data fit as well as efficient rank estimation without manual tuning. Finally, VB-PMF (and the other PMF-based approaches) outperforms BMF, showing the effectiveness of our approach (and low-rank PMF models in general) in recommendation tasks.}
\begin{table}[t]
\renewcommand{\arraystretch}{1.3}
\centering
\caption{Performance on the MovieLens \rev{10M} dataset \rev{considering the 100 most rated movies. Since rank selection for CTF3D and SQ was done in parallel, we present the worst-case (among the candidate ranks) average runtime. The runtimes for the \revFive{benchmark} approaches are negligible and are not shown.} \revOneB{Due to computational constraints in terms of memory and runtime, CTF-Full-ValErr and TN-ValErr were not considered in this experiment.}}
\label{table:1}
\begin{tabular}{|l|c|c|cll}
\cline{1-4}
Algorithm      & RMSE                & MAE                 &  \multicolumn{1}{c|}{\begin{tabular}[c]{@{}c@{}}Runtime\\ (min)\end{tabular}}&  \\ \cline{1-4}
VB-PMF         & \textbf{0.872 $\pm$ 0.008} & 0.665 $\pm$ 0.006 &  \multicolumn{1}{c|}{\textbf{72.44}} &  \\ \cline{1-4}
CTF3D-ValErr   & \textbf{0.871 $\pm$ 0.008} & \textbf{0.659 $\pm$ 0.006} &  \multicolumn{1}{c|}{737.58} &  \\           
SQ-AIC         & 0.899 $\pm$ 0.011 & 0.686 $\pm$ 0.008 &  \multicolumn{1}{c|}{344.55} &  \\
SQ-BIC         & 0.926 $\pm$ 0.008 & 0.708 $\pm$ 0.005 &  \multicolumn{1}{c|}{344.55} &  \\
SQ-DNML        & 0.926 $\pm$ 0.008 & 0.708 $\pm$ 0.005 &  \multicolumn{1}{c|}{346.09} &  \\ \cline{1-4}
BMF            & 0.935 $\pm$ 0.018 & 0.706 $\pm$ 0.010 &  &  \\
User average   & 0.989 $\pm$ 0.007 & 0.780 $\pm$ 0.005 &  &  \\
Movie average  & 1.121 $\pm$ 0.006 & 0.902 $\pm$ 0.004 &  &  \\
Global average & 1.015 $\pm$ 0.005 & 0.839 $\pm$ 0.004 &  &  \\ \cline{1-3}
\end{tabular}
\end{table}
%

\section{Conclusion}

%
%
%
%

\rev{\revFive{We} presented a novel Bayesian framework for joint parameter and model order (rank) estimation of low-rank probability mass function (PMF) tensors. Unlike existing tensor-based PMF estimation approaches, which require the rank to be heuristically pre-selected or determined through computationally expensive validation or information criteria, the proposed method infers the rank directly from the observed data during inference. This is accomplished through a Bayesian model that exploits Dirichlet priors to enforce probability simplex constraints and to promote sparsity in the latent components, thereby enabling irrelevant rank-one terms to be pruned without \revFive{\it{ad hoc}} thresholding or repeated training.

The resulting VB-PMF algorithm yields closed-form posterior updates with deterministic convergence and offers computational efficiency while eliminating the need for cross-validation. Experiments using synthetic data demonstrate that the proposed method converges to the true tensor rank as the number of samples increases, and achieves \revFour{a} higher estimation accuracy and \revFour{a} better robustness to missing data than likelihood-based and coupled tensor factorization approaches. Furthermore, experiments on real classification and recommendation datasets demonstrate its practical effectiveness. On these tasks, VB-PMF equals or exceeds the performance of classical models like random forests and biased matrix factorization while offering an interpretable and versatile model based on the low-rank components of the estimated joint PMF.
}
\appendix \label{appendix}
\section*{Derivation of the ELBO} 
We begin by expanding the first term on the right-hand side of \eqref{eq:elbo_1} into its constituent distributions. This gives (cf. \eqref{eq:joint_distrib})
\begin{equation} \label{eq:elbo_1st_term}
    \begin{aligned}[b]
        \rev{\mathbb{E}[\log p(\Y, \THETA)]} & = \\
        & \hspace{-5em} \mathbb{E}\Big[\log p \big(\Y \, \big| \, \Z, \{\A_n\}_{n=1}^N \big)\Big] +  \mathbb{E}\Big[\log p(\Z \, | \, \lambdab)\Big] \\
         & \hspace{-4em}  + \mathbb{E}\Big[\log p(\lambdab)\Big] + \sum_{n=1}^N \sum_{r=1}^R \mathbb{E}\Big[\log p(\a_{n,r})\Big],
    \end{aligned}
\end{equation}
where the explicit dependence of the expected values on the variational distributions $q(\cdot)$ has been dropped for brevity. Next, we substitute for each distribution and take the expected value of each parameter with respect to its variational distribution in turn. Thus \revThree{(cf. \eqref{eq:log_joint})}, 
\begin{align}
    &\hspace{-6em}\mathbb{E}\Big[\log p \big(\Y \, \big| \, \Z, \{\A_n\}_{n=1}^N \big)\Big]  +  \mathbb{E}\Big[\log p(\Z \, | \, \lambdab)\Big]  \notag \\
    &\hspace{-6em}= \sum_{t=1}^T  \sum_{r=1}^R \mathbb{E}[z_{r,t}] \Big(\mathbb{E}\big[\log \lambda_r \big] + \sum_{n=1}^N  \mathbb{E}\big[\log a_{n,r,y_{n,t}}\big] \Big) \\
    \mathbb{E}\Big[\log p(\lambdab)\Big]  &=  \log C(\alphab_{\lambda}) + (\alpha_{\lambda} - 1) \sum_{r=1}^R \mathbb{E}[\log \lambda_r] \\
    \mathbb{E}\Big[\log p(\a_{n,r})\Big]   &= \log C(\alphab_{n,r}) \notag \\
    &+ (\alpha_{n,r} - 1)\sum_{\revFive{i}=1}^{I_n} \mathbb{E}[\log a_{n,r,\revFive{i}}]
\end{align}
\revFour{Calculating} the expectations \revFour{via} \eqref{eq:expectations} and \eqref{eq:z_est} yields the first three terms of \eqref{eq:elbo_2}. Similarly, the second term on the right-hand side of \eqref{eq:elbo_1} is given by (cf. \eqref{eq:mean_field})
\begin{equation} \label{eq:elbo_2nd_term}
    \begin{aligned}[b]
        \mathbb{E}[\log q(\THETA)] & = \sum_{t=1}^T \mathbb{E}[\log q_z(\z_t)] + \mathbb{E}[\log q_{\lambda}(\lambdab)] \\
         & + \sum_{n=1}^N \sum_{r=1}^R \mathbb{E}[\log q_{n,r}(\a_{n,r})].
    \end{aligned} 
\end{equation}
Substituting for the variational distributions gives \revThree{(cf. \eqref{eq:q_z_2}, \eqref{eq:opt_q_lambda}, and \eqref{eq:opt_q_nr})}
\begin{equation}
   \mathbb{E}[\log q_z(\z_t)]  = \sum_{r=1}^R \mathbb{E}[z_{r,t}] \log \rho_{r,t}
\end{equation}
\begin{align}
    \mathbb{E}[\log q_{\lambda}(\lambdab)] & = \log C(\widetilde{\alphab}_{\lambda}) + (\widetilde{\alpha}_{\lambda} - 1) \mathbb{E}[\log \lambda_r] \\
    \mathbb{E}[\log q_{n,r}(\a_{n,r})]  & = \log C(\widetilde{\alphab}_{n,r}) \notag  \\
        &  + (\widetilde{\alpha}_{n,r} - 1)\sum_{\revFive{i}=1}^{I_n} \mathbb{E}[\log a_{n,r,\revFive{i}}] 
\end{align}
after which the expectations are \revFour{computed} via \eqref{eq:expectations} and \eqref{eq:z_est} to yield the last three terms of \eqref{eq:elbo_2}.

\bibliographystyle{utils/IEEEtran}
\bibliography{references/bayesianPMF}

\end{document}